\documentclass[manuscript,nonacm]{acmart}
\usepackage{amsmath}
\usepackage{graphicx} %
\usepackage{physics}
\usepackage{tikz,tikz-3dplot}
\usepackage{hyperref}
\usepackage{todonotes}
\setuptodonotes{inline}
\makeatletter
\renewcommand{\listoftodos}[1][\@todonotes@todolistname]{\@starttoc{tdo}{#1}}
\makeatother
\usepackage{subcaption}
\usepackage{xspace} 
\usetikzlibrary{arrows.meta, calc, backgrounds, shapes.geometric}

\definecolor{fslgreen}{HTML}{2A9D6F}
\definecolor{fsldarkgreen}{HTML}{1A6E44}
\definecolor{fslblue}{HTML}{4A7FC1}
\definecolor{fslorange}{HTML}{D97706}

\newcommand{\fsl}{FishSense Lite\xspace}
\AtBeginDocument{%
  \providecommand\BibTeX{{%
    \normalfont B\kern-0.5em{\scshape i\kern-0.25em b}\kern-0.8em\TeX}}}

\renewcommand\footnotetextcopyrightpermission[1]{}  %
\begin{document}

\title[\fsl]{\fsl: A Camera-Based, Single Laser Hardware Framework with Software System for In Situ Fish Length Measurement
}

\author{Christopher L. Crutchfield}
\email{ccrutchf@ucsd.edu}
\orcid{0000-0002-2011-721X}
\affiliation{%
  \institution{University of California, San Diego}
  \city{La Jolla}
  \state{California}
  \country{USA}
}

\author{Kyle Hu}
\email{kshu@ucsd.edu}
\orcid{0009-0006-4494-7257}
\affiliation{%
  \institution{University of California, San Diego}
  \city{La Jolla}
  \state{California}
  \country{USA}
}

\author{Nathan Hui}
\email{nthui@ucsd.edu}
\orcid{0000-0002-0780-9350}
\affiliation{%
  \institution{University of California, San Diego}
  \city{La Jolla}
  \state{California}
  \country{USA}
}

\author{Omair Qazi}
\email{omqazi@ucsd.edu}
\orcid{0009-0006-4363-2931}
\affiliation{%
  \institution{University of California, San Diego}
  \city{La Jolla}
  \state{California}
  \country{USA}
}

\author{Jennifer Loch}
\email{jen.loch@reef.org}
\orcid{0000-0002-8782-3098}
\affiliation{%
  \institution{Reef Environmental Education Foundation}
  \city{Key Largo}
  \state{Florida}
  \country{USA}
}

\author{Alli Candelmo}
\email{alli@reef.org}
\orcid{0000-0001-7624-8596}
\affiliation{%
  \institution{Reef Environmental Education Foundation}
  \city{Key Largo}
  \state{Florida}
  \country{USA}
}

\author{Brice X. Semmens}
\email{bsemmens@ucsd.edu}
\orcid{0000-0001-5663-9194}
\affiliation{%
  \institution{Scripps Institution of Oceanography}
  \city{La Jolla}
  \state{California}
  \country{USA}
}

\author{Curt Schurgers}
\email{cschurgers@ucsd.edu}
\orcid{0000-0002-9005-7881}
\affiliation{%
  \institution{University of California, San Diego}
  \city{La Jolla}
  \state{California}
  \country{USA}
}

\author{Ryan Kastner}
\email{kastner@ucsd.edu}
\orcid{0000-0001-9062-5570}
\affiliation{%
  \institution{University of California, San Diego}
  \city{La Jolla}
  \state{California}
  \country{USA}
}

\renewcommand{\shortauthors}{Crutchfield, et al.}

\begin{abstract}
Fish stock assessments need fish lengths paired with species from far more places than trained surveyors can reach.
Recreational divers already visit many of those places, but measuring a fish underwater has required trained observers or specialist equipment.
We present \fsl, which recovers fish length from a single photograph taken with a recreational dive camera, a dive laser pointer, and a 3D-printed mount.
The diver aims the laser at a fish and takes a picture; afterward, the dot's position in the image gives the range to the fish, and the range turns the fish's size in the image into a length.
A citizen-science program, trained by the system's builders, taught its own divers and collected lengths paired with species for 73 wild fish across seven open-water dives.
On seven wild fish also measured by an underwater stereo-imaging system, the two instruments agreed without detectable bias (mean difference $+0.8\%$).
On rigid targets of known length, \fsl recovered every target to within 1.3\%, and 99\% of 1,055 frames fell within a 15\% error budget.
From the error sources we measure, we derive a field protocol: calibrate against a dive slate on every dive, photograph fish within $15^\circ$ of perpendicular and beyond 1~m, and take at least ten frames of each.
\end{abstract}



\maketitle

\hypersetup{
  pdfauthor={Christopher L. Crutchfield, Kyle Hu, Nathan Hui, Omair Qazi,
             Jennifer Loch, Alli Candelmo, Brice X. Semmens, Curt Schurgers,
             Ryan Kastner},
  pdfsubject={},
  pdfcreator={LaTeX}
}

\section{Introduction}\label{sec:introduction}

Fish are a major source of protein for more than a third of the world's population~\cite{Guillen2018}. Overfishing reduces fish populations, disrupts marine ecosystems, and puts this critical food source at a significant risk~\cite{Hilborn2020}. Managing that risk means knowing whether a stock is replacing itself with young fish, which requires knowing how many fish there are, which species they are, and how they are distributed across age classes~\cite{Jennings2001}. %
The length of an individual fish supports estimates of its weight and its age, and the distribution of lengths within a species supports estimates of the population's age structure~\cite{JERALDS.2008, Jennings2001}. A photograph of a free-swimming fish carries the animal's identity, and researchers have used individual recognition across images to estimate abundance~\cite{Pettit2025}. A photograph does not record the fish's scale, however, since a small fish photographed up close can fill as much of the frame as a large one farther away.

Collecting accurate lengths at scale remains difficult even for high-quality programs such as the California Collaborative Fisheries Research Program (CCFRP). Volunteer anglers catch the fish, and trained scientists on deck identify the species, measure its length, and release it~\cite{Hamilton2021}. Landed fish are comparatively easy to measure, and phone-based tools already do so on deck~\cite{crutchfield2025fishsense}. Unfortunately, CCFRP is expensive and requires trained scientists on-site. Because of high assessment costs, many important areas, such as the Indo-Pacific, remain unassessed, creating blind spots in our understanding of the world's fish populations~\cite{Hilborn2020}. The RAM Legacy Stock Assessment Database, which compiles stock assessments worldwide, is concentrated in a few regions and omits seven of the ten largest wild-capture fishing nations~\cite{Hilborn2020, Ricard2011}. Without assessments, there is no basis for setting catch limits, and no way to know whether a fishery is being fished down until it has been.

Two things follow about collecting the missing data. First, scientist-staffed programs cannot easily reach enough places. The regions missing from the assessment databases are many and far apart, and staffing and transporting trained people to each of them is the cost that makes CCFRP hard to replicate. Scaling past that means removing the dependency on the scientist and letting the volunteers CCFRP already has work independently. Second, researchers should not catch fish to measure them. Bringing an animal to the surface exposes it to barotrauma, injury caused by the rapid drop in pressure as it rises~\cite{Rummer2005, Jarvis2008, Pribyl2011}, and to air~\cite{cook2015}. Measuring fish in situ avoids both, pointing to divers rather than anglers. Recreational divers are already in the water, including in the regions the databases miss. In the US alone, 2.5 million people are active divers and 11 million are snorkelers~\cite{DEMA2024}. Volunteer fish surveys are already collected during divers' regular dives~\cite{PattengillSemmens2003}, so divers go where scientists already want data. A diver who is already there, and who is not asked to catch anything, reaches places a scientist cannot and leaves the fish in the water.

In programs like CCFRP, volunteers contribute labor while a trained scientist is present and owns the measurement. Underwater, volunteers can instead collect observations without a scientist present, and scientists then work from what the volunteers bring back, an arrangement we refer to as independent citizen science. The Reef Environmental Education Foundation (REEF) Volunteer Fish Survey Project already works this way, training divers to survey reef fish on their own and record each species and its abundance~\cite{PattengillSemmens2003, Ashley2022}, and its data support published analyses of fish abundance and distribution~\cite{Ashley2022, Stallings2009, WardPaige2010}. Because it is not limited by the number of scientists available, independent citizen science can grow with the number of volunteers, although someone still needs to process measurements taken from images afterward. The volunteer survey records species and abundance, but not length.

The goal of this work is an instrument that lets a recreational diver measure fish on their own. Scientists already have technologies that measure fish underwater, including underwater stereo camera systems and laser calipers. The question is whether any of them can be adapted for recreational divers, or whether a different instrument is needed. Martin et al.~\cite{martin2016marine} found that marine volunteers judge a project's technical demands before joining, and that participants may drop out preemptively over anticipated difficulty. Each point at which a volunteer must buy something unusual, carry something extra, or hold expertise underwater is a point at which a volunteer may leave. We therefore set three design goals, each aimed at one of those points, for an instrument a volunteer diver uses without a scientist present.

\begin{enumerate}
    \item \textbf{Deferred judgment}: The volunteer is asked only to do what cannot be done later. Identifying the species and determining the length are judgments, and deferring them to people equipped to make them keeps the diver's burden to the one step that has to happen in the water.

    \item \textbf{Commodity components}: The system should use no specialized parts. Any component that must be specified, quoted, or sourced from a specialist supplier creates a stop point, so every part should be something a recreational diver already owns or can order from a consumer retailer.

    \item \textbf{Manual-only training}: The system should require no training beyond a written manual. Training that only a scientist can deliver reintroduces the same bottleneck as having a scientist on the dive, so the system should be usable from written instructions by someone who has not used it before.
\end{enumerate}

\begin{figure}
    \centering
    \rotatebox{270}{\includegraphics[scale=0.15]{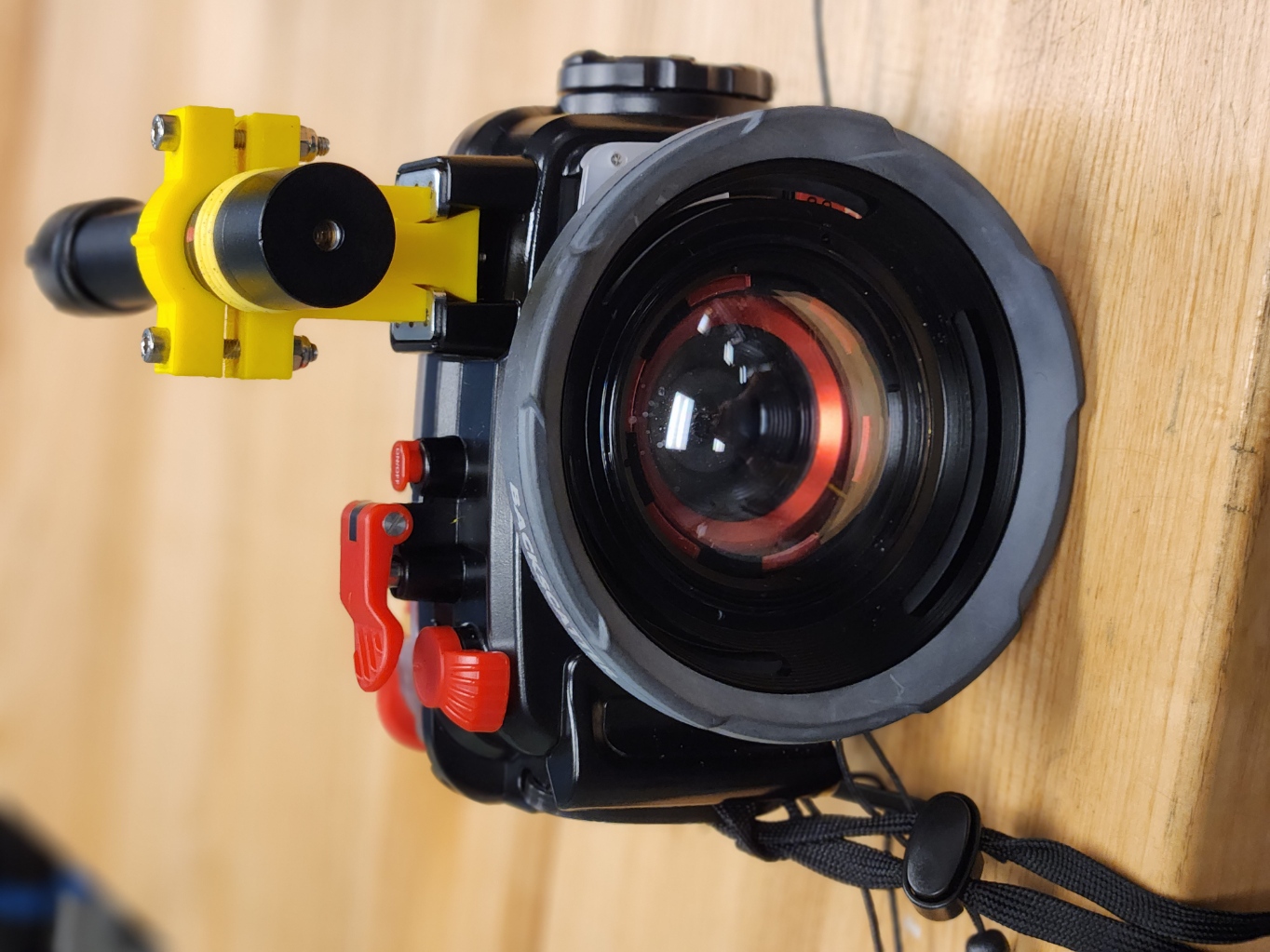}}
    \caption{The \fsl system adds a laser pointer to the Olympus TG-6 commercial-off-the-shelf (COTS) dive camera. A recreational diver takes a picture of a fish with the laser pointed at it. Our software calculates the fish length using only that information.}
    \Description{A photo of the \fsl System.}
    \label{fig:system}
\end{figure}

Figure~\ref{fig:system} shows \fsl, a commercial-off-the-shelf (COTS) dive camera such as the popular Olympus Tough TG-6, modified only by a laser dive pointer rigidly attached by a 3D-printed mount. The diver positions the laser point on the fish and takes a picture, and software recovers metric length from that single image. Their experience is akin to ``virtual spearfishing,'' where each laser tag on a fish is a potential measurement of fish length. The printed mount is the only part that is not bought, and the dive slate the system calibrates against is one most divers already carry. This paper argues that commodity hardware assembled this way measures fish well enough to add volunteer length data alongside existing visual surveys, rather than replace them. We hold it to a 15\% error budget per length, a design target set with the fisheries scientists we worked with, and in Section~\ref{sec:results} account for how that budget is spent.

Our primary contributions are:
\begin{enumerate}
    \item \textbf{A fish measurement system built from commodity components}: A handheld device combining a COTS dive camera and a dive laser pointer with a software pipeline that recovers metric fish length from a single image, calibrated against a dive slate.

    \item \textbf{A characterization of what that system measures}: An accounting of the system's accuracy against targets of known length and against a state-of-the-art stereo-imaging system, together with the sources of error that consume its 15\% budget.

    \item \textbf{Evidence that a citizen-science program can operate it without its builders:} We partnered with REEF, providing an initial training to its staff, who then trained their own volunteer divers. REEF’s staff and volunteers later collected the field measurements reported here.
\end{enumerate}

Section~\ref{sec:background} reviews existing approaches, Section~\ref{sec:system} describes \fsl, Section~\ref{sec:setup} describes the evaluation, Section~\ref{sec:results} presents its results, and Section~\ref{sec:discussion} discusses what they mean for the next design.

\section{Background}\label{sec:background}

Existing methods of collecting fish length each meet some of the three design goals, and laser calipers come closest, meeting all three in principle.

Catch and release is a common way to obtain fish length, but it is expensive at scale and harmful to the fish. Between 2007 and 2020, the California Collaborative Fisheries Research Program (CCFRP) relied on 1,700 volunteer anglers and the crews of 36 ships to catch 176,594 fish, averaging only 282 per day trip~\cite{Hamilton2021}. Catching the fish is itself invasive~\cite{Bartholomew2005}, through air exposure before release~\cite{COOKE2007, cook2015, Hanson2007} and barotrauma that can force a fish's stomach out through its mouth~\cite{Rankin2017, Rummer2005, Jarvis2008}, and it may kill the fish the project aims to preserve~\cite{Bartholomew2005, Pribyl2011}.

Underwater visual surveys avoid handling the animal entirely and can be done anywhere accessible by scuba divers. Volunteer surveys such as REEF's record species and abundance~\cite{PattengillSemmens2003, Ashley2022}. Length is recorded in scientific visual surveys, where trained divers estimate it by eye~\cite{Delacy2017, Harvey2001}, which requires divers to identify species and estimate length underwater. Both are judgments that \emph{deferred judgment} would move to after the dive, and the training that supports length estimation is lost without practice~\cite{Bell1985}, which is what \emph{manual-only training} is meant to avoid. These surveys work well for trained divers, and their length data depends on that training.

Instead of estimating length in the water, a diver can record the animal and have the measurement taken afterward, which is where the technologies below come in.

Underwater stereo cameras extract distance and length information from the disparity between two views~\cite{Gedge2011, Mallet2014, Goetze2019}. This approach is accurate and serves as the reference against which this paper measures performance in Section~\ref{sec:results}. Commercial underwater stereo systems require purchasing proprietary hardware and software~\cite{Goetze2019}. Conventional systems cost at least US\$6,200 for calibration equipment, software, and cameras~\cite{Delacy2017}. Low-cost rigs built from action cameras reduce the price~\cite{Delacy2017}, but they still require two cameras held rigidly on a base bar and synchronized at the start of each recording~\cite{Delacy2017, Goetze2019}, and any movement of the cameras disrupts the stereo calibration and introduces measurement error~\cite{Goetze2019}, so the pair must be recalibrated against a checkerboard whenever it shifts~\cite{Delacy2017}. That recurring procedure is what \emph{manual-only training} is meant to avoid, and it has kept stereo with trained scientific divers.

Commercial all-in-one depth cameras offer the same range measurement in a single device. Tueller et al. built a diver-held fish measurement system around an Intel RealSense depth camera behind a flat acrylic port~\cite{Tueller2021, Wong2022}, and Wong et al. demonstrated the complexity of calibrating such a camera underwater: calibrated in air, it read depth 25.9\% off, and Intel's own calibration tools could not bring both depth and length within acceptable error underwater~\cite{Wong2022}. The same system produced more data than its on-board computer could write at full frame rate~\cite{Tueller2021}. Its color camera also records only $1280 \times 800$ pixels~\cite{Tueller2021}, which leaves less detail for the species and individual identifications that \emph{deferred judgment} moves to after the dive. \fsl avoids the flat-port problems Wong et al.\ observed by correcting its own refraction in hardware (Section~\ref{sec:device_and_workflow}).

Acoustic cameras (imaging sonar) recover range well and have been used to observe fish where cameras cannot~\cite{Mueller2006}, but at roughly US\$80,000 an instrument in 2006~\cite{Mueller2006} they fall outside the \emph{commodity components} design goal, and their returns give a shape and a distance without the species a stock assessment pairs with length.

Researchers have attempted to adapt LiDAR for underwater use multiple times, and the most successful efforts have involved adding green lasers to the infrared used on land~\cite{Szafarczyk2022, Filisetti2018}. This works because green light penetrates much farther than red or infrared~\cite{pope1997absorption}. Even with these modifications to typical terrestrial LiDAR, their range is significantly reduced~\cite{Filisetti2018}. Such devices remain niche, with only a handful of commercial products~\cite{Davis2024}, which puts them out of reach for the same reason as stereo; like acoustics, they return geometry without appearance. A camera records what the fish looks like, and what it lacks is the distance these methods provide.

Other methods keep a single camera to record the animal and add a separate way to measure its distance. Recovering range from where a laser spot falls in a nearby camera image is a well-established technique~\cite{Barreto2013, Matos2020}, and a monocular variant instead tracks a handheld laser's pose in each image~\cite{Takatsuka1999, Takatsuka2003}. Underwater, a single laser pointer offset from a camera has given a vehicle's distance to a wall~\cite{Muljowidodo2009}, and laser-line triangulation has been calibrated behind flat~\cite{Chantler1997, Matos2020} and dome~\cite{Bjerkeng2021} ports. Laser triangulation has also been applied to fish length in aquaculture tanks, with line lasers on fixed camera rigs calibrated underwater against a marked plate~\cite{Li2024, Wang2026}. Jaquet sized sperm whales from a boat by pairing a camera with a handheld laser rangefinder and scaling each image's pixels by the measured range~\cite{Jaquet2006}, and Pilgrim et al.\ scaled ROV survey images of the seabed with five lasers on the camera: four parallel lasers set the image's perspective and scale, and a fifth, set at an angle, gave the range from where its spot fell~\cite{Pilgrim2000}. Both measure range for a camera rather than replacing it, the same division of labor \fsl uses. The underwater systems among these are mounted on vehicles or fixed rigs, whereas \fsl is held by a diver in the field. An earlier version of this system appears in Hu's master's thesis~\cite{Hu2024},
with initial pool tests on data separate from those reported here, and the
measurement method is described in a patent application~\cite{Kastner2025}.

Laser calipers are the closest existing approach divers use to measure fish, and in principle they meet all three design goals. These systems position two parallel lasers at a known distance and ask the diver to put both dots on a fish, with the distance between the dots serving as a reference length~\cite{Rohner2011, Heppell2012}. The camera records and the components are standard, so the diver estimates nothing. The method has supported long-term monitoring of fish size with minimal training~\cite{Heppell2012, Stock2021}. Its accuracy rests on the two lasers staying parallel at their known separation, so practitioners check them, though how often varies: once on land~\cite{Heppell2012}, regularly on land~\cite{Rohner2015}, underwater against a reference before measuring~\cite{Deakos2010}, before each dive~\cite{Stock2021}, or at the start and end of each encounter, with the encounter's data discarded when the spacing has drifted by the end~\cite{Wong2018}. The method therefore depends on keeping the lasers parallel and on landing both dots on the fish, which makes the laser baseline the shortest fish it can measure.

\begin{table*}[]
    \centering
    \begin{tabular}{|p{0.15\textwidth}||p{0.30\textwidth}|p{0.27\textwidth}|p{0.16\textwidth}|}
        \hline
        Technique & Barrier to a volunteer & Accuracy & Range\\
        \hline\hline
        Stereo Imagery & two calibrated cameras; recalibrated whenever the rig shifts~\cite{Goetze2019, Delacy2017} & $-0.6$~cm mean error~\cite{Harvey2001} & 2--10~m~\cite{Mallet2014} \\\hline
        Acoustic Methods & Specialist instrument~\cite{Mueller2006} & $\pm 0.5$~cm at 1~m to $\pm 6$~cm at 12~m~\cite{Mueller2006} & 1--12~m~\cite{Mueller2006} \\\hline
        Laser Caliper & Consumer components; lasers kept parallel~\cite{Heppell2012, Stock2021} & 5--7\% ($<10^\circ$ off axis)~\cite{Stock2021} & 0.5--5~m~\cite{Heppell2012, Stock2021} \\\hline
        \fsl (this paper) & Consumer components and a printed mount; slate photographed each dive & $-2.2\%$ median per frame (pool); 8.9\% vs.\ stereo (Section~\ref{sec:results}) & 0.5--5~m, beyond 1~m recommended (Section~\ref{sec:results}) \\\hline
    \end{tabular}
    \caption{Comparison of different underwater ranging techniques. Ranges are those reported or tested in the cited work.}
    \label{tab:comparison}
\end{table*}

Table~\ref{tab:comparison} collects these techniques. Among the methods divers use to measure fish, \fsl is closest to the laser caliper and differs by using a single laser whose pose is measured from photographs of a slate rather than set by alignment, so only one dot needs to land on the fish, and no laser spacing sets the shortest fish it can measure; half the field fish of Section~\ref{sec:field_results} were shorter than the 25~cm spacing of one caliper~\cite{Heppell2012}. Unlike the single-laser and fish-length systems above, which are mounted on vehicles or fixed rigs and calibrated before deployment, \fsl is diver-held, recalibrated on every dive in the field from a slate at an unknown pose, and evaluated in a citizen-science deployment.

\section{\fsl}\label{sec:system}

\begin{table*}[t]
    \centering
    \begin{tabular}{|l|l|l|r|}
        \hline
        Component & Source & Typically owned? & Share of cost \\
        \hline\hline
        Dive camera & Consumer retailer & Yes, by underwater photographers & 45\% \\\hline
        Underwater housing & Consumer retailer & Yes, with the camera & 31\% \\\hline
        Wide-angle wet lens & Consumer retailer & Sometimes & 16\% \\\hline
        Dive laser pointer & Dive retailer & Sometimes & 7\% \\\hline
        Laser mount & 3D printed from files released with the paper & No, printed & 1\% \\\hline
    \end{tabular}
    \caption{Components of the \fsl system. Every part is available through consumer channels, and none requires a quote or a specialist supplier. The mount is the only part that must be made rather than bought. Shares are of October 2026 retail prices for an equivalent unit, about US\$1,300.}
    \label{tab:system_components}
\end{table*}

\fsl is a handheld device built from COTS components, paired with software that recovers metric fish length from a single photograph. The diver photographs the fish, and the system computes the length afterward. This section describes the ideas that make a single-image measurement possible, the device that implements them (Section~\ref{sec:device_and_workflow}), the geometry that recovers a length (Section~\ref{sec:length_recovery}), the calibration that geometry depends on (Section~\ref{sec:calibration}), and the assumptions the whole chain rests on (Section~\ref{sec:assumptions}).

\subsection{Core design ideas}
\label{sec:design_ideas}

\begin{figure*}[t]
\centering

\begin{subfigure}[b]{0.30\textwidth}
\centering
\resizebox{\linewidth}{!}{%
\begin{tikzpicture}[
    >={Latex[length=2mm, width=1.5mm]},
    font=\large,
]
    \fill[fslblue, opacity=0.35] (-0.7, -1.8) rectangle (0.7, 0);
    \draw[fslblue, line width=0.7pt] (-0.7, -1.8) rectangle (0.7, 0);
    \node[gray, font=\small, anchor=north] at (0, -1.85) {camera};

    \fill[fslblue, opacity=0.07]
        (0, 0) -- (-3, 7.5) -- (3, 7.5) -- cycle;
    \draw[fslblue, opacity=0.55, line width=0.4pt] (0, 0) -- (-3, 7.5);
    \draw[fslblue, opacity=0.55, line width=0.4pt] (0, 0) -- (3, 7.5);

    \fill[fslblue, opacity=0.07]
        (0, 0) -- (-0.72, -1.8) -- (0.72, -1.8) -- cycle;
    \draw[fslblue, opacity=0.55, line width=0.4pt] (0, 0) -- (-0.72, -1.8);
    \draw[fslblue, opacity=0.55, line width=0.4pt] (0, 0) -- (0.72, -1.8);

    \draw[gray!80, line width=1pt] (-0.7, -1.8) -- (0.7, -1.8);
    \node[gray, font=\small\itshape, anchor=west]
        at (0.78, -1.8) {image plane};

    \draw[fslorange, line width=0.6pt, dash pattern=on 1.5pt off 1.5pt]
        (2, 6.8) -- (0, 0) -- (-0.529, -1.8);
    \draw[fslorange, line width=0.6pt, dash pattern=on 1.5pt off 1.5pt]
        (-2, 6.8) -- (0, 0) -- (0.529, -1.8);

    \fill (0, 0) circle (1.5pt);
    \node[font=\large\bfseries, anchor=south west]
        at (0.1, 0.05) {O};

    \draw[fslgreen, line width=1.8pt, line cap=round]
        (-2, 6.8) -- (2, 6.8);
    \node[fslgreen, font=\large\bfseries, anchor=west]
        at (2.1, 6.8) {L};
    \node[gray, font=\small\itshape, anchor=south]
        at (0, 7) {fish};

    \draw[fslgreen, line width=1.8pt, line cap=round]
        (-0.529, -1.8) -- (0.529, -1.8);
    \node[fslgreen, font=\large\bfseries, anchor=north west]
        at (0.6, -1.95) {p};

    \draw[<->, line width=0.5pt] (-1.4, 0) -- (-1.4, -1.8);
    \node[font=\large\bfseries, anchor=east] at (-1.5, -0.9) {f};

    \draw[<->, line width=0.5pt] (-3.4, 0) -- (-3.4, 6.8);
    \node[font=\large\bfseries, anchor=east] at (-3.5, 3.4) {d};
\end{tikzpicture}}
\caption{Pinhole camera}
\label{fig:principle:pinhole}
\end{subfigure}
\hfill
\begin{subfigure}[b]{0.28\textwidth}
\centering
\resizebox{\linewidth}{!}{%
\begin{tikzpicture}[
    >={Latex[length=2mm, width=1.5mm]},
    font=\large,
]
    \fill[fslblue, opacity=0.35] (-0.7, -1.8) rectangle (0.7, 0);
    \draw[fslblue, line width=0.7pt] (-0.7, -1.8) rectangle (0.7, 0);
    \node[gray, font=\small, anchor=north] at (0, -1.85) {camera};

    \fill[fslblue, opacity=0.35] (0.78, -1.8) rectangle (1.18, -0.27);
    \draw[fslblue, line width=0.7pt] (0.78, -1.8) rectangle (1.18, -0.27);
    \node[gray, font=\small, anchor=north]
        at (0.98, -1.85) {laser};

    \fill[fslblue, opacity=0.07]
        (0, 0) -- (-3, 7.5) -- (3, 7.5) -- cycle;
    \draw[fslblue, opacity=0.55, line width=0.4pt] (0, 0) -- (-3, 7.5);
    \draw[fslblue, opacity=0.55, line width=0.4pt] (0, 0) -- (3, 7.5);

    \draw[fslgreen, line width=0.9pt] (0.98, -0.27) -- (0.98, 7.5);

    \fill[fslgreen] (0.98, 3.2) circle (3pt);
    \node[font=\large\bfseries, anchor=east] at (0.85, 3.2) {A};
    \fill[fslgreen] (0.98, 5.5) circle (3pt);
    \node[font=\large\bfseries, anchor=east] at (0.85, 5.5) {B};
\end{tikzpicture}}
\caption{Camera with laser}
\label{fig:principle:laser}
\end{subfigure}
\hfill
\begin{subfigure}[b]{0.30\textwidth}
\centering
\resizebox{\linewidth}{!}{%
\begin{tikzpicture}[font=\footnotesize]
    \draw[gray!80, line width=0.8pt] (0,0) rectangle (4,3);
    \node[gray, font=\scriptsize\itshape, anchor=north west] at (0,-0.05) {image};

    \coordinate (vp) at (2,1.5);
    \coordinate (far) at (1.45,3);
    \draw[fslgreen, line width=0.9pt, dash pattern=on 3pt off 2pt] (far) -- (vp);

    \fill[fslgreen] ($(vp)!0.75!(far)$) circle (2.4pt);
    \node[font=\footnotesize\bfseries, anchor=east] at ($(vp)!0.75!(far)+(-0.08,0)$) {A};
    \fill[fslgreen] ($(vp)!0.35!(far)$) circle (2.4pt);
    \node[font=\footnotesize\bfseries, anchor=east] at ($(vp)!0.35!(far)+(-0.08,0)$) {B};

    \draw[line width=0.9pt] ($(vp)+(-0.09,-0.09)$) -- ($(vp)+(0.09,0.09)$);
    \draw[line width=0.9pt] ($(vp)+(-0.09,0.09)$) -- ($(vp)+(0.09,-0.09)$);
    \node[font=\scriptsize, anchor=west] at ($(vp)+(0.15,-0.05)$) {vanishing point};

    \draw[->, gray, line width=0.5pt] ($(vp)!0.95!(far)+(0.35,0)$) -- ($(vp)!0.15!(far)+(0.35,0)$)
        node[midway, right, font=\scriptsize\itshape, gray] {farther};
\end{tikzpicture}}
\caption{What the camera sees}
\label{fig:principle:image_plane}
\end{subfigure}

\caption{The principle of laser triangulation underlying \fsl. \textbf{(a)} A pinhole camera projects a fish of length $L$ at distance $d$ to an inverted image of length $p$. \textbf{(b)} A laser mounted offset from the camera; points A and B are at different distances along the laser ray. \textbf{(c)} The same points in the image, sliding toward the beam's vanishing point as distance grows.}
\label{fig:principle}
\end{figure*}
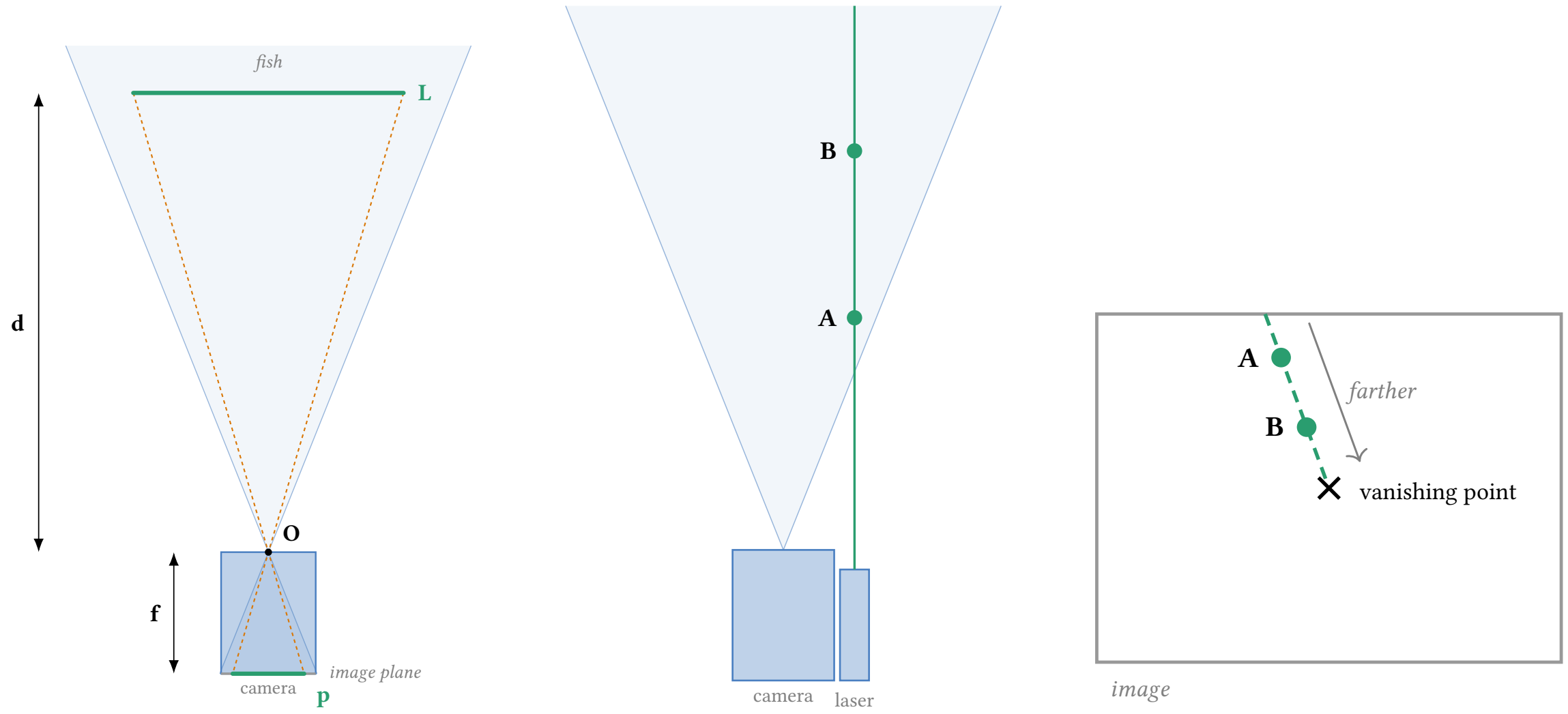

Given a photograph of a fish, we want its length. A camera answers part of this immediately, since the fish covers some number of pixels in the image, which we call its \textit{pixel length}. Pixel length alone does not give a metric length, because a small fish close to the camera can cover the same number of pixels as a large fish farther away, so a given pixel length is consistent with many different fish. Knowing the distance to the fish resolves this ambiguity. As illustrated in Figure~\ref{fig:principle}\,(a), a fish of true length $L$ at distance $d$ from the camera produces a pixel length $p$, and by similar triangles $p / f = L / d$, where $f$ is the camera's focal length in pixels. Rearranged, $L = p \cdot d / f$, a first-order relation that Section~\ref{sec:length_recovery} replaces with an exact back-projection. The image gives $p$, and the camera calibration gives $f$, leaving only $d$. For now we take the fish to lie broadside to the camera, an assumption Section~\ref{sec:assumptions} lists and Section~\ref{sec:pose_results} tests.

\begin{figure}
    \centering
    \includegraphics[width=\columnwidth]{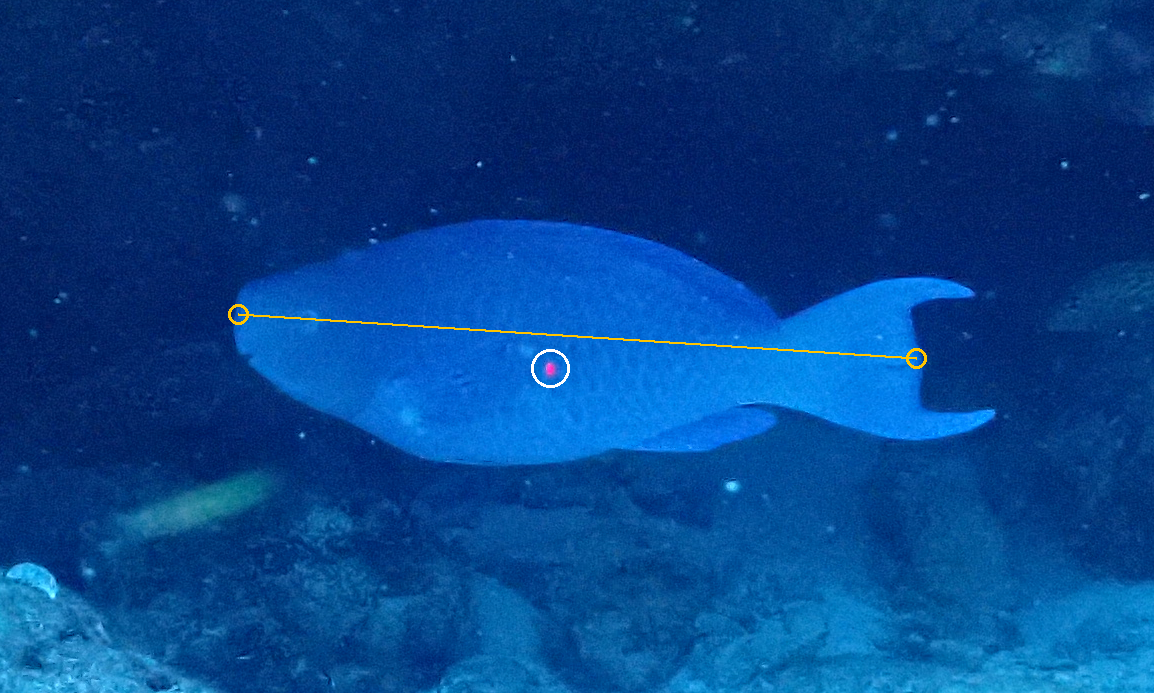}
    \caption{A \fsl frame. The white ring marks the laser dot; its position in the image gives the fish's range. The amber marks are the snout and tail fork as the labeler placed them, and the length is the distance between those two points back-projected to the dot's depth.}
    \Description{An underwater photograph of a blue parrotfish with the laser dot and the two measured endpoints marked.}
    \label{fig:parrotfish}
\end{figure}

\fsl recovers $d$ with a laser pointer mounted rigidly beside the camera (Figure~\ref{fig:principle}\,(b)). When the laser strikes an object, the dot appears somewhere in the image. Move that object farther away, and the dot moves across the frame, because the camera sees the beam from the side. The dot's position therefore indicates how far away the surface is. Figure~\ref{fig:principle}\,(c) shows the path it traces. As distance grows, the dot converges toward a fixed point in the image, the vanishing point of the beam, and closer distances put the dot farther from that point. This is \textit{laser triangulation}, which has been used to measure the range from an ROV's camera to the seabed alongside parallel lasers that set the image scale~\cite{Pilgrim2000}. In \fsl, a single laser provides both range and scale, so no parallel lasers are needed; one photograph of the fish contains everything the measurement needs. We measure the laser's position and direction from photographs of a dive slate during each dive (Section~\ref{sec:calibration}). Figure~\ref{fig:parrotfish} shows a frame of a fish: the dot gives the range, and the two endpoints define the pixel length.

\subsection{Device and Workflow}
\label{sec:device_and_workflow}

\fsl consists of a COTS underwater camera with a waterproof laser pointer attached by a custom 3D-printed mount, as shown in Figure~\ref{fig:system}. Our reference implementation uses an Olympus Tough TG-6 in its standard PT-059 underwater housing, with a Backscatter M52 Underwater 81$^\circ$ Wide Air Lens mounted at the housing's port. Behind a flat port, light bends where it passes from the water into the air inside the housing, so the camera is not strictly a single-viewpoint pinhole underwater, and a pinhole model approximates it well only at a constant range~\cite{treibitz2012flat}. Divers measure fish across a range of distances, so that approximation does not hold for \fsl. The M52 restores the pinhole behavior assumed by the model in Section~\ref{sec:length_recovery} at every range, without refraction-specific calibration, while keeping the camera's topside field of view. The laser is a Class IIIA waterproof dive pointer, limited to 5~mW of visible output~\cite{CFR1040}. Divers used both red and green pointers, and because the geometry below depends only on where the dot lands in the image, it is the same for either.

Table~\ref{tab:system_components} lists the system components. The mount clamps the laser to the housing's cold shoe, offset from the camera's optical axis, through a printed taper, body, and cap. The mount is the only part that cannot be bought off the shelf, but it requires no specialist manufacturing.
We will release the design files on publication, along with the analysis code. A diver can print the mount in polylactic acid (PLA) on a 3D printer, or order it from an online print service like any other part, so the mount stays within the \emph{commodity components} design goal. An equivalent unit costs about US\$1,300 at October 2026 retail prices, three-quarters of it in the camera and housing an underwater photographer already owns (Table~\ref{tab:system_components}).

To measure a fish's length, the diver aims the laser at the fish's body and holds the shutter button for a short burst of pictures, recorded as RAW stills rather than video for the reason given in Section~\ref{sec:length_recovery}. A usable frame has the laser dot on the fish and both the snout and the tail fork visible. The laser dot also serves as a framing aid: because the beam is nearly parallel to the optical axis, keeping the dot on the fish keeps the fish near the center of the image. Section~\ref{sec:instructions} provides the complete set of instructions divers follow.

The system needs two calibrations. The camera intrinsics are fitted once per camera, underwater against a checkerboard, and do not change. The laser extrinsics are fitted for every dive from photographs of a dive slate the diver already carries, and Section~\ref{sec:calibration} describes the procedure. After the dive, images are transferred to a computer, where human annotators label the laser dot, snout, and tail-fork pixel locations. \fsl's software pipeline then extracts a metric length for each pictured fish. This division follows the \emph{deferred judgment} design goal, leaving the diver only the photography that has to happen in the water. Automating the labeling step is a data analysis problem rather than a system one, and it is out of scope for this paper.

\subsection{Length Recovery}
\label{sec:length_recovery}

We define \textit{fish length} as fork length: the distance from the snout to the tail fork. This section turns the snout and fork pixels a labeler marks into that physical length. The similar triangles in Section~\ref{sec:design_ideas} assumed a known distance and a fish centered in the image; here, the distance comes from the laser, and the endpoints may lie anywhere in the frame. Recovery requires the laser's position and direction in the camera frame, which are not known in advance and are recovered by the calibration procedure in Section~\ref{sec:calibration}.

We work in a coordinate system with the origin at the camera's focal point and the $z$-axis aligned with the optical axis. We model the camera with the basic pinhole camera model~\cite{hartley2003multiple}, in which a 3D point $\mathbf{P}$ expressed in this frame projects to a pixel location $\mathbf{q}$. A pixel location fixes a direction from the camera and not a distance, so the projection holds only up to an unknown positive scale. Writing that scale explicitly as $\lambda_1$,
\begin{equation}
    \label{eq:pinhole}
    \lambda_1 \mathbf{q} = K \mathbf{P},
\end{equation}
where $K$, the camera intrinsic matrix, contains the focal lengths $f_x$ and $f_y$ in pixels along each image axis and the principal point $(c_x, c_y)$, the pixel where the optical axis meets the image. Standard intrinsic calibration~\cite{Zhang2000, bradski2008learning} recovers it once per camera, underwater, using a checkerboard. We fit that calibration to the camera's RAW output, and we use RAW for the measurement for the same reason. The camera's JPEG and video modes apply factory processing that $K$ does not describe, so a length recovered from either would be using a calibration for a different optical path. Rearranging Equation~\ref{eq:pinhole} maps each pixel back to its original 3D coordinate,
\begin{equation}
    \label{eq:camera_projection}
    \mathbf{P} = \lambda_1 K^{-1} \mathbf{q}.
\end{equation}
In an observed image, however, $\lambda_1$ is an unknown scale factor that must be recovered, so a pixel alone fixes only a ray through the camera center along which the original 3D point lies.

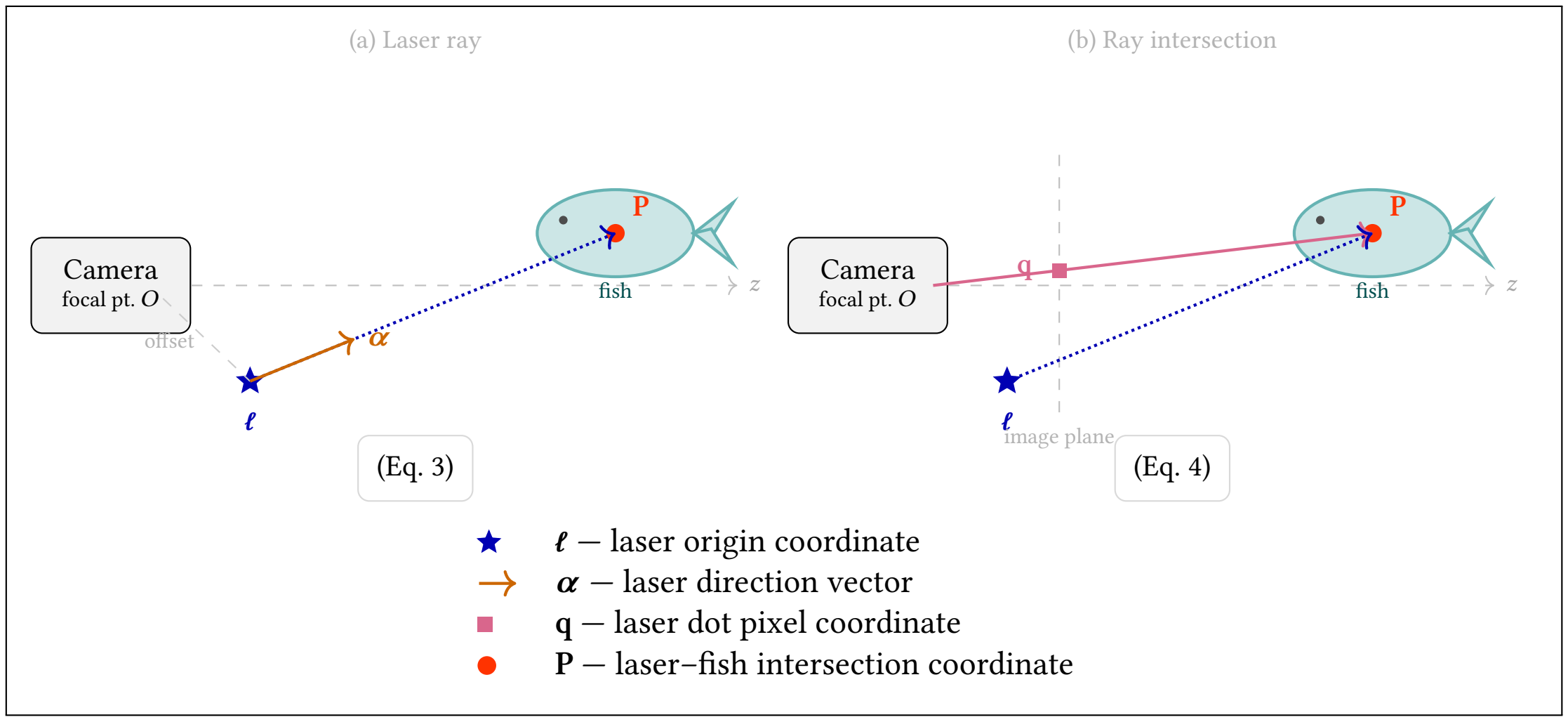
\begin{figure*}[t]
    \centering
    \fbox{%
        \begin{minipage}{0.95\textwidth}
            \centering
            \begin{tikzpicture}[
                scale=0.82,
                every node/.style={font=\small},
                ray/.style={->, thick},
                axis/.style={->, dashed, gray!50, thin},
                leader/.style={dashed, gray!40, thin},
                lstar/.style={star, star points=5, star point ratio=2.2,
                              fill=blue!70!black, draw=blue!70!black,
                              inner sep=0pt, minimum size=7pt},
            ]

            \node[gray!60, font=\footnotesize] at (3.5, 2.8) {(a) Laser ray};
            \node[draw, rounded corners=3pt, fill=gray!10,
                  minimum width=1.5cm, minimum height=0.9cm] (cam) at (0,0)
                  {\shortstack{Camera \\ \scriptsize focal pt.\ $O$}};
            \draw[axis] (cam.east) -- (7.2, 0)
                node[right, gray!70, font=\scriptsize] {$z$};
            \draw[fill=teal!20, draw=teal!60, thick]
                (5.8, 0.6) ellipse (0.9cm and 0.5cm);
            \draw[fill=teal!20, draw=teal!60, thick]
                (6.7, 0.6) -- (7.15, 0.95) -- (6.9, 0.55) -- cycle;
            \draw[fill=teal!20, draw=teal!60, thick]
                (6.7, 0.6) -- (7.15, 0.25) -- (6.9, 0.65) -- cycle;
            \fill[black!70] (5.2, 0.75) circle (1.4pt);
            \node[below=10pt, teal!60!black, font=\scriptsize] at (5.8, 0.6)
                {fish};
            \coordinate (ell) at (1.6, -1.1);
            \node[lstar] at (ell) {};
            \node[below=5pt, blue!70!black] at (ell) {$\boldsymbol{\ell}$};
            \draw[leader] (0.6,-0.15) -- (ell)
                node[midway, left, gray!60, font=\scriptsize] {offset};
            \coordinate (P) at (5.8, 0.6);
            \fill[red!60!orange] (P) circle (3pt);
            \node[above right=2pt, red!60!orange] at (P) {$\mathbf{P}$};
            \draw[->, thick, blue!70!black, densely dotted]
                (ell) -- (P);
            \draw[->, thick, orange!80!black]
                (ell) -- ++(1.2, 0.486)
                node[right, orange!80!black] {$\boldsymbol{\alpha}$};
            \node[draw=gray!30, rounded corners=3pt,
                  fill=white, inner sep=5pt,
                  font=\small] at (3.5, -2.1)
                {(Eq.~\ref{eq:laser_ray})};

            \begin{scope}[xshift=8.7cm]

            \node[gray!60, font=\footnotesize] at (3.5, 2.8)
                {(b) Ray intersection};
            \node[draw, rounded corners=3pt, fill=gray!10,
                  minimum width=1.5cm, minimum height=0.9cm]
                  (cam2) at (0,0)
                  {\shortstack{Camera \\ \scriptsize focal pt.\ $O$}};
            \draw[axis] (cam2.east) -- (7.2, 0)
                node[right, gray!70, font=\scriptsize] {$z$};
            \draw[fill=teal!20, draw=teal!60, thick]
                (5.8, 0.6) ellipse (0.9cm and 0.5cm);
            \draw[fill=teal!20, draw=teal!60, thick]
                (6.7, 0.6) -- (7.15, 0.95) -- (6.9, 0.55) -- cycle;
            \draw[fill=teal!20, draw=teal!60, thick]
                (6.7, 0.6) -- (7.15, 0.25) -- (6.9, 0.65) -- cycle;
            \fill[black!70] (5.2, 0.75) circle (1.4pt);
            \node[below=10pt, teal!60!black, font=\scriptsize] at (5.8, 0.6)
                {fish};
            \coordinate (ell2) at (1.6, -1.1);
            \node[lstar] at (ell2) {};
            \node[below=5pt, blue!70!black] at (ell2)
                {$\boldsymbol{\ell}$};
            \coordinate (P2) at (5.8, 0.6);
            \fill[red!60!orange] (P2) circle (3pt);
            \node[above right=2pt, red!60!orange] at (P2) {$\mathbf{P}$};
            \draw[dashed, gray!50] (2.2, 1.5) -- (2.2, -1.5)
                node[below, gray!60, font=\scriptsize] {image plane};
            \coordinate (q) at (2.2, 0.172);
            \fill[purple!60]
                ($(q)+(-2.3pt,-2.3pt)$) rectangle ($(q)+(2.3pt,2.3pt)$);
            \node[left=4pt, purple!60] at (q) {$\mathbf{q}$};

            \draw[->, thick, purple!60]
                (0.75, 0) -- (P2);
            \draw[->, thick, blue!70!black, densely dotted]
                (ell2) -- (P2);
            \node[draw=gray!30, rounded corners=3pt,
                  fill=white, inner sep=5pt,
                  font=\small] at (3.5, -2.1)
                {(Eq.~\ref{eq:setup})};
            \end{scope}

            \end{tikzpicture}

            \vspace{0.5em}
            \begin{tabular}{ll}
                \tikz[baseline=-0.6ex]{\node[star, star points=5,
                    star point ratio=2.2, fill=blue!70!black,
                    draw=blue!70!black, inner sep=0pt,
                    minimum size=6pt]{};} &
                    $\boldsymbol{\ell}$ --- laser origin coordinate \\
                \tikz[baseline=-0.6ex]{\draw[->, thick, orange!80!black]
                    (0,0) -- (0.35,0);} &
                    $\boldsymbol{\alpha}$ --- laser direction vector \\
                \tikz[baseline=-0.6ex]{\fill[purple!60]
                    (-2pt,-2pt) rectangle (2pt,2pt);} &
                    $\mathbf{q}$ --- laser dot pixel coordinate \\
                \tikz[baseline=-0.6ex]{\fill[red!60!orange]
                     (0,0) circle (2.5pt);} &
                    $\mathbf{P}$ --- laser--fish intersection coordinate \\
            \end{tabular}
            \vspace{0.5em}

        \end{minipage}
    }
    \caption{Laser ray model and ray intersection. \textbf{(a)} The laser, with origin $\boldsymbol{\ell}$ and direction $\boldsymbol{\alpha}$, intersects the fish at $\mathbf{P}$. \textbf{(b)} The camera ray and laser ray share the terminus $\mathbf{P}$, forming an overdetermined system in $\lambda_1$ and $\lambda_2$.}
    \label{fig:ray_model}
\end{figure*}

The laser supplies a second equation for the same point. We define the laser origin as $\boldsymbol{\ell}$ and the laser direction as $\boldsymbol{\alpha}$, the laser's two extrinsic parameters in the camera frame, both illustrated in Figure~\ref{fig:ray_model}. The point at which the laser intersects the fish is
\begin{equation}
    \label{eq:laser_ray}
    \mathbf{P} = \lambda_2 \boldsymbol{\alpha} + \boldsymbol{\ell},
\end{equation}
where $\lambda_2$ parameterizes distance along the laser ray. The dot we see in the image is where these two rays meet, so setting Equation~\ref{eq:camera_projection} equal to Equation~\ref{eq:laser_ray} gives
\begin{equation}
    \label{eq:setup}
    \lambda_1 K^{-1} \mathbf{q} = \lambda_2 \boldsymbol{\alpha} + \boldsymbol{\ell}.
\end{equation}
This system is overdetermined, with three scalar equations in two scalar unknowns, so we solve for $\lambda_1$ by linear least squares. Let $\mathbf{r} = K^{-1}\mathbf{q}$ denote the back-projected camera ray. Projecting the residual of Equation~\ref{eq:setup} onto $\mathbf{r}$ and onto $\boldsymbol{\alpha}$ and eliminating $\lambda_2$ yields
\begin{equation}
    \label{eq:lambda2}
    \lambda_1 =
    \frac{\boldsymbol{\ell}^\top \mathbf{r}
          - (\boldsymbol{\alpha}^\top \boldsymbol{\ell})
            (\boldsymbol{\alpha}^\top \mathbf{r})}
         {\|\mathbf{r}\|^2
          - (\boldsymbol{\alpha}^\top \mathbf{r})^2}.
\end{equation}
The denominator vanishes when $\boldsymbol{\alpha}$ and $\mathbf{r}$ are parallel, which the camera ray approaches as the dot recedes toward the beam's vanishing point, so the scatter in range from imprecisely placed labels, which we call \textit{triangulation noise}, grows with range (Section~\ref{sec:range_results}). The baseline between the laser and the camera sets how quickly that happens: a wider baseline keeps the two rays further from parallel at a given distance, and so extends the range over which depth is well determined. The reconstructed 3D point is $\mathbf{P} = \lambda_1 \mathbf{r}$, and since $r_z = 1$ we have $P_z = \lambda_1$ directly. We use $\lambda_1$ as the fish's metric depth.

Unlike the similar-triangles shortcut of Section~\ref{sec:design_ideas}, which is exact only at the image center and requires $f_x = f_y$, a condition that is not guaranteed~\cite{hartley2003multiple, bradski2008learning}, we back-project the snout and tail fork pixel locations through $K^{-1}$ and place each at the depth the laser gave us:
\begin{equation}
    \label{eq:snout}
    \mathbf{P}_{\text{snout}} = P_z \cdot K^{-1} \mathbf{q}_{\text{snout}}
\end{equation}
\begin{equation}
    \label{eq:tail}
    \mathbf{P}_{\text{tail}} = P_z \cdot K^{-1} \mathbf{q}_{\text{tail}}
\end{equation}
The metric fish length is the Euclidean distance between the back-projected points:
\begin{equation}
    \label{eq:fish_length}
    L = \| \mathbf{P}_{\text{snout}} - \mathbf{P}_{\text{tail}} \|.
\end{equation}
Handling the endpoints this way keeps the measurement exact for a fish anywhere in the frame and for sensors whose pixels are not square, which the similar-triangles shortcut does not. It recovers the fish's length projected onto a plane at the laser's depth. When the fish lies broadside to the camera, that projection is the fish's length. When it does not, the projected length is shorter, an effect called \textit{foreshortening}, which Section~\ref{sec:pose_results} measures as its own error term. The laser's depth is itself that of the fish's near flank, closer to the camera than the line from snout to fork, so a thick fish also reads short, which Section~\ref{sec:pool_results} measures as \textit{body depth}.

\paragraph{From frames to a length.}
Length comes from several frames of the same fish rather than one, because pose can only shorten a frame's measurement. A fish angled away from the camera, or bent mid-stroke, measures short, so the frame taken closest to broadside gives the longest measurement. Other errors, such as a misplaced endpoint or noise in the dot's position, can push a single frame either way, so we report the $p_{90}$ of an individual's measurements rather than their maximum. Under the nearest-rank definition we use, an individual with fewer than ten frames is assigned its longest frame, which leaves such an estimate exposed to a single frame that reads long; in the pool, estimates from two to eight frames instead read slightly short on average, by 1.4\% at two frames (Section~\ref{sec:pool_results}). Section~\ref{sec:pool_results} also shows that the $p_{75}$ and $p_{90}$ bracket the known length, while the median reads short and the maximum long.

\subsection{Calibration}
\label{sec:calibration}

Every length depends on where the laser points, and on \fsl, which changes from one dive to the next. Equation~\ref{eq:lambda2} takes the laser's origin $\boldsymbol{\ell}$ and direction $\boldsymbol{\alpha}$ as given, so error in either becomes error in depth, and through depth, error in length. The mount is printed plastic registered to a cold shoe, and the beam it holds shifts between dives by a median of $0.65^\circ$. Measuring one dive's fish with another dive's calibration raises the median length error from 3.1\% to 25.2\% (Section~\ref{sec:calibration_results}). The laser's pose is therefore measured anew on every dive.

\begin{figure}[h]
    \centering
    \includegraphics[
        angle=180,
        trim={190pt 40pt 40pt 70pt},
        clip,
        width=0.5\columnwidth
    ]{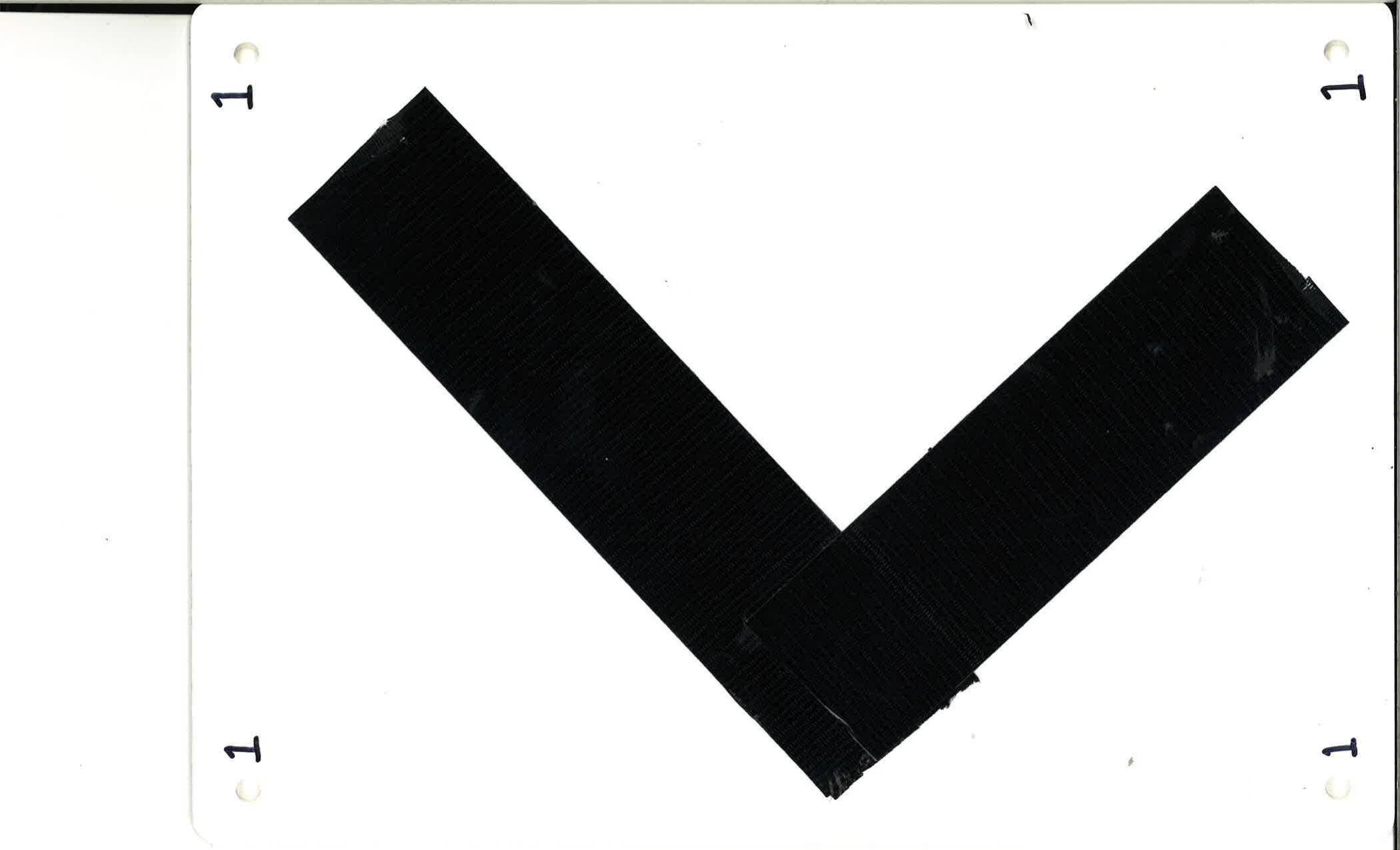}
    \caption{Field calibration target: a standard underwater dive slate with a chevron of black duct tape.}
    \label{fig:dive_slate}
\end{figure}

The calibration recovers the beam's pose from images of the dot landing on a flat object of known geometry, and it assumes only that the object is flat and its pattern known, not that it is held square to the camera. Any planar target with labeled features at known spacing will do. A printed camera-calibration checkerboard is the standard choice, and our earliest pool sessions used one, but in the field we use a standard dive slate marked with duct tape, shown in Figure~\ref{fig:dive_slate}, because a slate is already in the diver's kit and asks nothing extra of them. At the start or end of a dive, a buddy holds the slate while the distance between the slate and the camera is swept outward until the laser becomes hard to keep on the slate, then back in, and the diver photographs the slate whenever the dot is on it. The procedure follows the slate check used with some laser calipers~\cite{Stock2021}. Divers aim for about ten frames, more than the fit needs, because some will be unusable. Sweeping in range is what fixes the laser's direction, and Section~\ref{sec:calibration_results} tests how much range the sweep needs.

For each calibration image, passing the labeled slate corners and the pattern's known metric geometry to OpenCV's \texttt{solvePnP} routine~\cite{OpenCVsolvePnP}, using its default iterative solver, recovers a rotation $R_i$ and translation $\mathbf{t}_i$ mapping the slate's body frame to the camera frame. The slate plane $\Pi_i$ follows, with normal $\mathbf{n}_i$ the rotated slate body-frame $z$-axis, and any slate corner transformed to the camera frame, $\mathbf{p}_i$, giving a point on the plane. Intersecting the camera ray through the labeled laser dot pixel $\mathbf{q}_i$ with $\Pi_i$ recovers the laser dot's 3D position:
\begin{equation}
    \label{eq:ray_plane_intersection}
    \mathbf{P}_i =
    \frac{\mathbf{n}_i^\top \mathbf{p}_i}
         {\mathbf{n}_i^\top K^{-1} \mathbf{q}_i}
    \cdot K^{-1} \mathbf{q}_i.
\end{equation}
This produces observed 3D points $\{\mathbf{P}_i\}_{i=1}^n$, each lying, up to labeling noise, on the laser ray. The laser direction $\boldsymbol{\alpha}$ is the normalized sum of sign-corrected pairwise differences:
\begin{equation}
    \label{eq:alpha_fit}
    \boldsymbol{\alpha} =
    \frac{\mathbf{d}}{\|\mathbf{d}\|},
    \qquad
    \mathbf{d} = \sum_{i \ne j}
        \operatorname{sgn}\!\bigl((P_i - P_j)_z\bigr)
        \,(\mathbf{P}_i - \mathbf{P}_j),
\end{equation}
and the laser origin $\boldsymbol{\ell}$ is the point at which the fitted ray intersects the $z=0$ plane:
\begin{equation}
    \label{eq:ell_fit}
    \boldsymbol{\ell} =
    \bar{\mathbf{P}} -
    \frac{\bar{P}_z}{\alpha_z}\, \boldsymbol{\alpha},
\end{equation}
where $\bar{\mathbf{P}}$ is the mean of the $\mathbf{P}_i$. Equation~\ref{eq:alpha_fit} is a least-squares fit of the beam's lateral position against the rank of each point's range, so widely spaced points, which best fix the direction, count most. When the frames are evenly spaced in range, it equals regressing lateral position on depth. It differs from a total-least-squares fit only by a term that grows with the beam's tilt from the optical axis, and on the field calibrations the two directions agree to within $0.03^\circ$. These two parameters complete Equation~\ref{eq:lambda2}, and together they yield a length from a single labeled frame.

\subsection{Assumptions}
\label{sec:assumptions}

The chain from photograph to length rests on a small number of assumptions, which fall into two kinds.

Two are preconditions, in that the measurement does not exist without them. The laser must produce a single point, and that point must land on the fish. A dot on the reef behind the animal returns the reef's depth instead of the fish's.

The rest are approximations, in that the measurement degrades in a way we can characterize. We treat the camera as a pinhole, which the corrective optic in Section~\ref{sec:device_and_workflow} makes valid underwater at every range. We treat the fish as flat, perpendicular to the camera axis, and rigid, so both endpoints lie at the laser's depth and neither moves between them. The laser's pose is treated as fixed for the duration of a dive. Each holds only approximately, and Section~\ref{sec:results} tests them: the fixed laser pose in Section~\ref{sec:calibration_results}, the perpendicular fish in Section~\ref{sec:pose_results}, the flat fish in the body-depth comparison of Section~\ref{sec:pool_results}, and the pinhole model end-to-end through the accuracy on targets of known length in the same section.

\section{Evaluation Setup}\label{sec:setup}

The evaluation draws on three cohorts that differ in what they can be scored against, all collected with the same instrument and instructions.

\subsection{What the diver is asked to do}
\label{sec:instructions}

Divers learned the system from REEF rather than from the group that built it.
We trained the program's staff in a series of meetings using a written manual covering assembly, camera settings, the field procedure, and guidelines for taking usable photographs; the staff then trained their divers in an in-person briefing and ran the deployments.

The guidelines reduce to six rules, and each protects something the measurement depends on.

\begin{enumerate}
    \item \textbf{Point the laser at the fish, as close to its center of body mass as possible.} A dot on the reef behind the animal returns the reef's depth instead of the fish's.

    \item \textbf{Do not use the zoom.} Zooming changes the focal length, and $K$ was fitted with no zoom.

    \item \textbf{Do not touch or move the laser, and do not use it as a handle.} The calibration holds only while $\boldsymbol{\ell}$ and $\boldsymbol{\alpha}$ hold.

    \item \textbf{Hold the camera's optical axis at approximately a right angle to the fish.} This is the perpendicularity assumption, and Section~\ref{sec:results} reports what deviation costs.

    \item \textbf{Keep the fish between 0.5 and 5 meters away, ideally around 3 meters.} Section~\ref{sec:results} revises the near end of this range.

    \item \textbf{Make sure the fish is illuminated.} The annotator has to find the snout and the tail fork.
\end{enumerate}

The manual also asks divers to photograph the slate at the start and end of each dive, following the procedure of Section~\ref{sec:calibration}.
None of the rules requires a certification.

\subsection{The deployment}
\label{sec:deployment}

We provided \fsl units to REEF, whose staff and volunteer divers had not used the instrument before.
Between August 2023 and January 2025, they collected nearly 70,000 images, and this paper's results rest on nearly 2,700 of those that became length measurements, since producing a measurement requires a person to label each frame.

This paper evaluates the instrument on three cohorts, summarized in Table~\ref{tab:cohorts}.
They differ in what they can be scored against.
The paired stereo dives have a second instrument measuring the same animals, the pool cohort has targets of known length, and the field cohort has neither.

\begin{table}[]
    \centering
    \begin{tabular}{|l|r|r|l|}
        \hline
        Cohort & Frames & Targets & Reference \\
        \hline\hline
        Paired stereo dives & 39 & 7 wild fish & stereo-imaging system \\\hline
        Pool & 1,055 & 5 rigid models & tape or measuring board \\\hline
        Field & 162 & 73 wild fish & none \\\hline
    \end{tabular}
    \caption{The three cohorts this paper evaluates. The angle experiment's 1,428 frames are reported separately (Section~\ref{sec:pose_results}).}
    \label{tab:cohorts}
\end{table}

\subsection{Paired stereo dives}
\label{sec:stereo_cohort}

The paired stereo dives test the system on live fish in open water, the condition it is built for, against a second instrument that measured the same fish at the same time.
Stereo imagery measures length with a mean error of $-0.6$~cm on targets of known length~\cite{Harvey2001}, so it serves as the reference here, though not as ground truth.
The stereo-imaging system was a diver-operated SeaGIS stereo camera system built on two GoPro cameras, calibrated against a SeaGIS calibration cube with the CAL software at least once per field trip, and its lengths were measured in EventMeasure~\cite{SeaGISHardware}.
A \fsl unit was carried alongside the stereo-imaging system on the same dives, and the two instruments captured seven fish in common, matched by species, markings, background, and the time each recorded them.

\subsection{Pool cohort}
\label{sec:pool_cohort}

The pool cohort establishes what the instrument measures when the answer is known in advance, which is what lets its error sources be separated and measured one at a time.
Over several days in 2023, the engineering team and REEF's staff photographed five rigid targets in three pools, using several cameras, housings, and laser units.
Table~\ref{tab:pool_targets} lists the targets.
Four are rigid plastic fish models, and one is a box; we measured each once with a tape or measuring board to determine the length the system is scored against.
Two of the fish models are flat, and two have the body depth of a real fish, which Section~\ref{sec:pool_results} uses to measure body-depth cost.

\begin{table}[]
    \centering
    \begin{tabular}{|l|r|l|}
        \hline
        Target & Length & Profile \\
        \hline\hline
        Box & 150 mm & flat \\\hline
        Purple Angelfish & 192 mm & flat \\\hline
        Rainbow Trout & 313 mm & thick \\\hline
        Grouper & 360 mm & flat \\\hline
        Snook & 455 mm & thick \\\hline
    \end{tabular}
    \caption{Rigid targets used in the pool. The flat targets match the planar assumption of Section~\ref{sec:assumptions}, while the thick targets carry the body depth of a real fish.}
    \label{tab:pool_targets}
\end{table}

A diver held the targets rather than fixing them to a rig, so pose varied at the holder's discretion.
The targets were photographed from 0.25 to 5.5~m away, with a median of 2~m, covering the manual's 0.5 to 5~m working range and extending beyond it at both ends.
The lasers were calibrated against a checkerboard in the earlier sessions and against the dive slate in the later ones.
Some pool sessions were measured with a calibration taken earlier, as the protocol then allowed.
The cohort comprises the sessions whose calibration was judged to fit by checking that a rigid object reads the same length at every range, a test that needs only a fixed size, not a known one, and fails when the 95\% interval of the slope of length against range lies entirely beyond $\pm 2\%$ per meter; the calibrations that failed it no longer described the laser, which is why calibration moved to every dive.
Section~\ref{sec:pose_results} reports the five sessions of the angle experiment separately.

\subsection{Field cohort}
\label{sec:field_cohort}

The field cohort is a realistic deployment of \fsl.
Seven open-water dives in the Florida Keys, between November 2023 and December 2024, yielded 162 measurements of 73 wild fish.
This cohort has no reference instrument and no known length, so it shows how consistently the system measures wild fish in use, not how accurately it measures them.
Of the seven dives, three were calibrated at the start, two at the end, and two at both.
Every field calibration spanned at least 0.9~m of range, above the 0.6~m Section~\ref{sec:calibration_results} requires.

\section{Results}\label{sec:results}

\subsection{Comparison with a stereo-imaging system}
\label{sec:stereo_results}

\begin{figure}
    \centering
    \includegraphics[width=0.7\columnwidth]{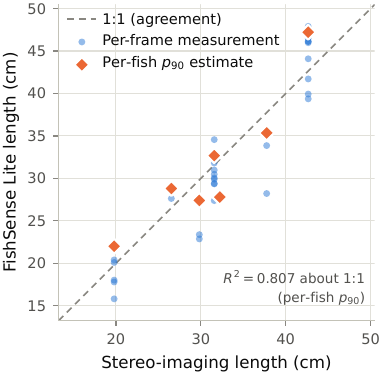}
    \caption{\fsl against an underwater stereo-imaging system on the same seven individuals. Each orange diamond is one fish's \fsl length, the $p_{90}$ of its frames; the blue points are those frames, 39 in all, between 3 and 11 per fish. The dashed line marks agreement, not truth; neither axis is a known length.}
    \Description{Scatter plot of \fsl length against stereo-imaging length for seven fish, with a 1:1 line.}
    \label{fig:paired_vs_stereo}
\end{figure}

We first test \fsl against stereo imagery on an open-water dive, where the two instruments, pointed at the same wild fish, show no detectable bias across the seven fish compared.
We measured the wild fish ranging from 20 to 43~cm with both instruments on one day (Figure~\ref{fig:paired_vs_stereo}), using each \fsl length as the $p_{90}$ of that fish's frames.
The estimates fall on the $1:1$ line with $R^2 = 0.807$ and a residual RMS of 3.0~cm.

Individual fish differ by up to 14\% (Table~\ref{tab:paired_fish}).
The mean difference is $+0.8\%$ (95\% confidence interval $-8.6$ to $+10.2\%$), the mean absolute difference is 8.9\%, and the 95\% limits of agreement~\cite{BlandAltman1986} run from $-19.1\%$ to $+20.7\%$.
With seven fish, those limits are themselves uncertain by about $\pm$16 points (95\% confidence intervals $-35.3$ to $-2.8\%$ and $+4.4$ to $+36.9\%$).
Each of the seven fish lies within the 15\% budget, but seven fish are too few to say how far a single measurement can stray, so we infer per-measurement accuracy from known-length targets (Section~\ref{sec:pool_results}).
The difference also shows no detectable trend with fish size (95\% confidence interval $-0.15$ to $+0.16\%$ per millimeter) or with the number of frames per fish (Spearman $\rho = 0.57$, $p = 0.18$).

\begin{table}[]
    \centering
    \begin{tabular}{|l|r|r|r|}
        \hline
        Species & Dive & Frames & Difference from stereo \\
        \hline\hline
        Hogfish & 1 & 11 & $+3.5\%$ \\\hline
        Hogfish & 2 & 3 & $+8.6\%$ \\\hline
        Hogfish & 1 & 3 & $-13.8\%$ \\\hline
        Blue Parrotfish & 1 & 10 & $+10.7\%$ \\\hline
        Stoplight Parrotfish & 1 & 6 & $+11.2\%$ \\\hline
        Gray Snapper & 2 & 3 & $-8.1\%$ \\\hline
        Blue Parrotfish & 2 & 3 & $-6.4\%$ \\\hline
    \end{tabular}
    \caption{The seven fish measured by both instruments, from two dives under one morning calibration. Each difference is the \fsl $p_{90}$ minus the stereo length, as a percentage of the stereo length. Within each dive the differences span most of the full range (dive 1: $-13.8$ to $+11.2\%$; dive 2: $-8.1$ to $+8.6\%$), while the two dive means differ by 4.9 points.}
    \label{tab:paired_fish}
\end{table}

Part of the difference between the two instruments comes from sources the pool cannot reproduce.
Four fish read long, by up to 11.2\%, and pose only ever shortens a length.
The difference is not mainly frame-to-frame noise, since the stereo fish's own frames vary by a median of 6.9\%, and that scatter accounts for under half of the variance even if none of it is pose; each fish carries an offset between the two instruments.
Nor is it calibration, since the seven fish came from two dives under one morning calibration, and a calibration error shifts every fish on a dive alike, while the spread lies among fish on the same dive rather than between the two dives.
Likely contributors include the stereo system's own error, endpoints placed on a live fish that flexes between frames, and poses beyond $15^\circ$.

\subsection{Comparison with known lengths}
\label{sec:pool_results}

\begin{figure}
    \centering
    \includegraphics[width=0.7\columnwidth]{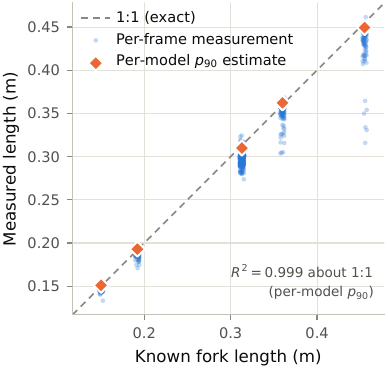}
    \caption{Measured against known length over the pool cohort, 1,055 frames of five rigid targets. Diamonds are the per-target $p_{90}$. $R^2 = 0.999$ about the $1:1$ line on those five estimates, and 0.976 on the individual frames.}
    \Description{Scatter plot of measured length against known length for five rigid targets.}
    \label{fig:measured_vs_known}
\end{figure}

Targets of known length let the instrument's own error be measured, rather than compared with another instrument's.
Against them, \fsl recovers each target to within 1.3\% at the $p_{90}$, and 99\% of individual frames fall inside the 15\% budget.
The per-target $p_{90}$ is $+0.65\%$ for the Box, $+0.47\%$ for the Purple Angelfish, $-1.00\%$ for the Rainbow Trout, $+0.59\%$ for the Grouper, and $-1.24\%$ for the Snook, with no trend across a three-fold length range (Figure~\ref{fig:measured_vs_known}).
A single frame reads a median of 2.2\% short, and 80\% of frames fall within 5\% of the known length and 98\% within 10\%.

\begin{figure}
    \centering
    \includegraphics[width=0.7\columnwidth]{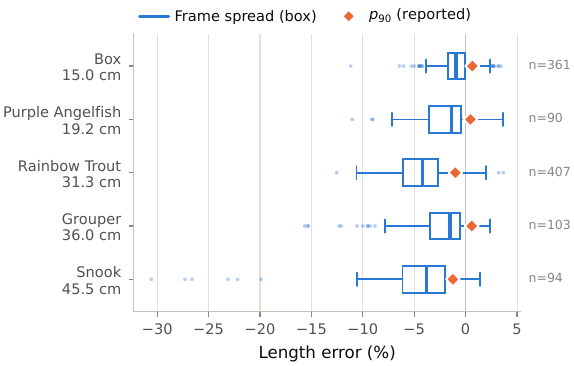}
    \caption{Percent length error by target. Box and whiskers give the frame-level spread, the diamond the reported $p_{90}$. All 1,055 frames are shown, out to the worst at $-30.6\%$.}
    \Description{Box plot of percent length error for each of the five rigid targets.}
    \label{fig:error_by_target}
\end{figure}

The error is one-sided, so we report a high quantile rather than a mean (Figure~\ref{fig:error_by_target}).
Each target's distribution has a long tail toward short readings and a short one toward long, as foreshortening requires, so a mean describes how divers held the targets rather than what the instrument measures.
The nine frames outside the budget all read short, and all come from two targets.
Their shortfall matches the foreshortening of a $32^\circ$ to $46^\circ$ pose, a cost Section~\ref{sec:pose_results} measures against a known angle.

\begin{figure}
    \centering
    \includegraphics[width=0.7\columnwidth]{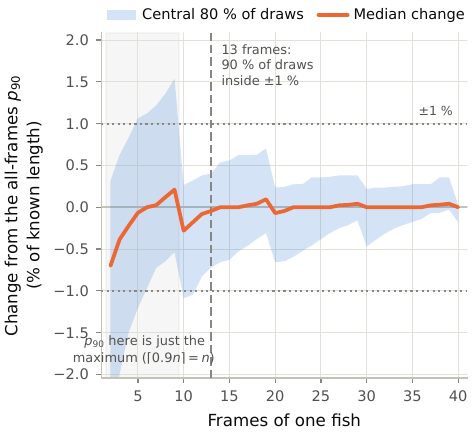}
    \caption{How far a pool target's $p_{90}$ moves when estimated from $n$ randomly drawn frames rather than all its frames, across the 15 sessions in which the Box or the Rainbow Trout was photographed at least 30 times. The shaded band holds 80\% of draws. Below ten frames, the $p_{90}$ is the longest frame.}
    \Description{Line plot of the change in the p90 estimate against the number of frames per target, with an 80 percent band of random draws.}
    \label{fig:p90_vs_frames}
\end{figure}

The estimate needs few frames, and its accuracy holds at the frame counts divers produce in the field.
Drawn at random from the 15 sessions in which the Box or the Rainbow Trout was photographed at least 30 times, the $p_{90}$ of four frames lands within 2\% of the all-frames value 90\% of the time, of ten frames within 1.5\%, and of thirteen within 1\% (Figure~\ref{fig:p90_vs_frames}).
Scored against the known length instead, the $p_{90}$ for two frames is 1.4\% short, and for eight frames within 0.1\%, since fewer than ten frames can only raise it.
The $p_{90}$ is also a balanced choice of quantile: across the five targets the median reads 1.5\% short and the maximum 3.4\% long, while the $p_{75}$ and $p_{90}$ bracket the known length.
We therefore recommend photographing each fish at least ten times, though fewer frames still give a usable length.

Thick models read slightly shorter than flat ones, and body depth (Section~\ref{sec:length_recovery}) explains part of the difference.
Under each of the seven checkerboard calibrations shared by the trout and the box, the trout read 3.0 points lower on average (95\% confidence interval 1.4 to 4.5).
In each of the five sessions that photographed the snook alongside the angelfish and the grouper, the snook read 1.8 points lower on average (95\% confidence interval 0.8 to 2.8).
For the trout, half its 58.7~mm body depth predicts a 1.7\% shortfall at its median range of 1.7~m, about half its observed gap.

\subsection{Laser calibration stability}
\label{sec:calibration_results}

\begin{figure}
    \centering
    \includegraphics[width=3.13in]{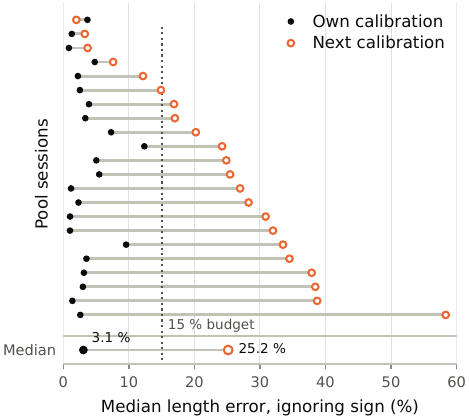}
    \caption{Each row is one pool session measured with its own calibration and, on the same frames, with the same unit's next calibration. A neighboring calibration moves 21 of 22 sessions further from their targets' known lengths and 16 of them past the 15\% budget, by an amount that tracks how far the beam moved between the two calibrations.}
    \Description{Dumbbell plot of each pool session's median length error with its own calibration and with the next calibration of the same unit, sorted by the latter.}
    \label{fig:calibration_substitution}
\end{figure}

Every length assumes the laser calibration is current, but a calibration records the laser's position and direction only when it was taken.
If the laser moves afterward, the measurement frames reveal some of that movement and hide the rest.
The camera's center and the laser beam together define a plane, which the camera views edge-on, the way a sheet of glass seen edge-on appears as a single line, so the dots the beam makes all fall along one line in the image.
A beam that tilts out of the plane moves the dots off that line, and the measurement frames show it.
A beam that rotates within the plane keeps the dots on the line.
The reprojection error, the distance between where the calibration predicts each dot and where it appears, is therefore unchanged.
Each position along the line now corresponds to a different range, though, so every length is rescaled.
Only a calibration can see this rotation, because the slate's known geometry gives each dot a range independent of the laser.
Recalibrating on every dive is therefore worth its cost if the laser rotates between dives by enough to matter, which the rest of this section measures.

Between dives, the beam rotates within the plane by more than the error budget allows.
At 2~m, a rotation of about half a degree ($-0.53^\circ$ to $+0.39^\circ$) puts a length outside the 15\% budget; between successive calibrations of a unit, the median rotation is $0.65^\circ$, exceeding that tolerance in 13 of 18 cases.
Between the pool tests and the 2024 deployments, three of the four units we can check rotated by about $3^\circ$.
The laser moves far less within a dive, where no new calibration is taken.
On two field dives, the slate was photographed at both the start and the end of the dive.
The calibration from one end matched the ranges measured at the other to within 3.7\%, well inside the 15\% budget.

Substituting one session's calibration for another's confirms that the laser moved, not that the fits were noisy.
If a calibration still described the laser at the next session, using it there would leave the error unchanged.
Instead, re-measuring each pool session with the same unit's next calibration raises the median error from 3.1\% to 25.2\% and puts 16 of 22 sessions outside the budget (Figure~\ref{fig:calibration_substitution}).
A borrowed calibration costs what the beam moved between the two calibrations (Spearman $\rho = 0.83$ between a session's error and that rotation), so calibration has to happen on every dive.

The change between calibrations comes from the laser, not from the calibration target.
A unit calibrated against both should report the same baseline, the laser's offset from the camera's optical center, and across the six units calibrated against both the checkerboard and the slate, the baselines differ by $+0.66\%$ (95\% confidence interval $-0.23\%$ to $+1.69\%$), within the variation between repeat calibrations against one target.

The movement comes from the joints in the mount, which stacks a cold shoe, a taper, the mount, a cap, and the pointer (Section~\ref{sec:device_and_workflow}).
Flex in the printed mount tilts the beam up or down, roughly within the plane because the laser sits above the lens; a twist at the cold shoe or the taper's screwed joint turns it sideways, mostly out of the plane; and the pointer turning in its clamp moves the beam around a small cone, because the laser diode is not exactly on the pointer's axis.
Between calibrations, the beam moves both ways but more in-plane than out of plane, with a median of $0.65^\circ$ in-plane vs. $0.35^\circ$ out-of-plane, and the in-plane component is larger in 11 of 18 steps, though a single step can be almost entirely one or the other.
Over deployments, the printed parts also crept, cracked, and absorbed water; screws pulled through them; and a published laser-caliper mount has likewise been reported to shift with temperature~\cite{Webster2010}.
An alignment mark on the pointer only detects the pointer turning in its clamp, so recalibrating at the dive is the only step that accounts for every joint.

The largest steps within a session came in the pool, across pauses in recording while the camera was out of use.
In one, the beam rotated $0.85^\circ$ within the plane between two consecutive calibrations.
In the other, the beam rotated $0.29^\circ$ partway through a session, after which that session's trout read 9.6\% short instead of 4.4\%.
A field diver, by contrast, holds one camera throughout the dive, without such pauses, and the two field dives calibrated at both ends held to within 3.7\% from start to finish, so on the dives we could check, one calibration per dive was enough.

A calibration also needs slate photographs taken across a range of distances.
Fitting the beam is like drawing a line through points: points close together let the line pivot, while points far apart hold it in place.
Mislocating the dot by one pixel at range $z$ moves it sideways by $z/f$, where $f$ is the focal length in pixels, and across a spread $\Delta z$ between the nearest and farthest photographs that error tilts the fitted beam by about $(z/f)/\Delta z$, so halving the spread doubles the tilt.
In one pool session, all the calibration photographs were taken within 3~cm of the same distance, so the calibration pinned the beam down only in that narrow neighborhood.
It reproduces ranges there to within 0.12\%, but reads 17\% short at a distance 2.2~m farther away.
None of the usual checks catches this, because the session's known-length targets sat near the calibration distance and measured well at $-2.4\%$, reprojection error is blind to in-plane rotation, and leaving out any one photograph leaves the same narrow neighborhood.
The distance between the slate and the camera is therefore swept outward and back during calibration (Section~\ref{sec:calibration}), and we require the calibration photographs to span at least 0.6~m of range.

\subsection{Fish pose}
\label{sec:pose_results}

\begin{figure}
    \centering
    \includegraphics[width=0.7\columnwidth]{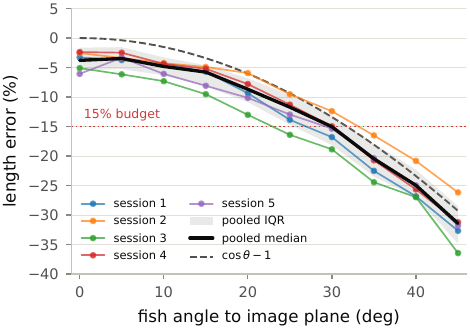}
    \caption{Percent length error against the fish's angle from perpendicular, from five sessions of one target stepped through $0^\circ$ to $45^\circ$. Thin lines are per-session medians, black is the pooled median and interquartile range, dashed is $\cos\theta - 1$, and dotted is the 15\% budget.}
    \Description{Line plot of percent length error against fish angle for five sessions.}
    \label{fig:error_vs_angle}
\end{figure}

Pose is the error source a diver controls most directly, and it costs length in a way the geometry predicts.
A fish angled away from the camera, so that its broadside is not perpendicular to the camera's axis, appears shorter than it is.
A fish angled $\theta$ from perpendicular projects to $\cos\theta$ of its length, as in Section~\ref{sec:length_recovery}.
Because \fsl places the snout and the fork at the laser's depth, it measures this projected length and reads $\cos\theta - 1$ short.
Setting that shortfall equal to an allowable error $\epsilon$ gives the largest angle a measurement can tolerate,
\begin{equation}
    \label{eq:foreshortening_theta}
    \theta = \cos^{-1}(1 - \epsilon),
\end{equation}
which is $31.8^\circ$ at the 15\% budget, and within $15^\circ$ of perpendicular foreshortening costs at most 3.4\%.

The pool data follow this curve over five sessions in which a Snook model was presented from $0^\circ$ to $45^\circ$ in $5^\circ$ steps, with the angle read off a protractor card held in frame, on five camera units at about 2 and 4~m, for 1,428 frames (Figure~\ref{fig:error_vs_angle}).
The sessions agree at every angle and in both ranges, and the pooled median follows the shape of $\cos\theta - 1$, crossing the budget at $30^\circ$ versus the predicted $31.8^\circ$.

The pooled curve sits below $\cos\theta - 1$ at every angle, by 3.8 points at $0^\circ$ and 2.4 at $15^\circ$; an offset pose cannot produce this, since a protractor card misaligned enough to explain the $0^\circ$ reading would need to be $15.9^\circ$ out, and would then put $15^\circ$ at $-14.2\%$ rather than the $-5.8\%$ observed.
Part of the offset is the Snook model's body depth, which reads it about 1.8 points below the flat models measured alongside it (Section~\ref{sec:pool_results}).
The rest is calibration, since a scale error fits the data best and differs by session, from $-6.5\%$ to $+0.2\%$ across the five, a spread that body depth, a wrong reference length, or a consistent habit in where labelers click could not produce, as each would shift all five sessions by similar amounts.
Of the $-5.8\%$ at $15^\circ$, $\cos 15^\circ - 1 = -3.4$ points is pose.

Within $15^\circ$ of perpendicular, every frame stays within budget, with room to spare.
Over the 542 frames inside $15^\circ$, the median error is $-4.7\%$, the $p_{90}$ is $-0.8\%$, and the worst is $-10.7\%$.
The first frames outside the budget appear at $25^\circ$, and the pooled median does not cross until $30^\circ$.
We therefore recommend photographing fish within $15^\circ$ of perpendicular, which keeps foreshortening small enough to leave room in the budget for the other sources of error.
Laser-caliper studies draw a similar line, reporting errors of 5--7\% within $10^\circ$ of perpendicular and 10--20\% beyond~\cite{Stock2021}, and rejecting frames past $10^\circ$~\cite{Heppell2012}.

\subsection{Working range}
\label{sec:range_results}

\begin{figure}
    \centering
    \includegraphics[width=0.7\columnwidth]{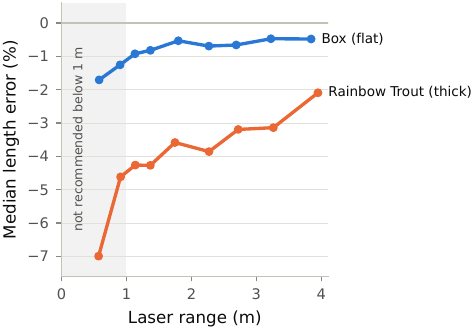}
    \caption{Median length error against laser range for the Rainbow Trout, a thick model 58.7~mm deep, and the Box, a flat face, in fixed range bins over the pool cohort. Shaded: ranges under 1~m.}
    \Description{Line plot of median length error against laser range for the trout and the box, with ranges under 1 m shaded.}
    \label{fig:working_range}
\end{figure}

Range decides how far from a fish a diver can work, and across most of it, length error does not grow with distance.
Beyond 0.82~m the median error of the pool cohort stays between $-2.4\%$ and $-1.5\%$, and its spread, measured as the interquartile range, stays between 3.0 and 4.7 points.
This holds out to 5~m, although we took only 30 frames beyond 4~m.
The geometry does not set that limit, since at 5~m the usual error in placing the dot shifts the range by only 1.5\%.
In practice, beyond about 5~m the dot became hard both to see on a fish and to keep on it.

Triangulation noise (Section~\ref{sec:length_recovery}) does grow with range, but too slowly to show in these data.
From the typical error in placing the labels, 0.9~px for the dot and 2.3~px for the snout and fork, it stays inside the observed spread out to about 5~m and would reach the 15\% budget only beyond 18~m.

Close to the camera, length error depends on the target rather than on the range (Figure~\ref{fig:working_range}).
Below 0.82~m, the trout reads $-7.0\%$ and the box only $-1.7\%$.
That gap is body depth (Section~\ref{sec:length_recovery}), because the laser lands on the trout's near flank, a fixed distance in front of the snout and fork, so the shortfall grows as a fraction of the range as the fish comes closer, and it predicts 5.0\% at those frames' median range of 0.58~m.
Close in, the gap between the two targets matches that prediction, though farther out the trout still reads about two points shorter than depth alone gives, from a source the data do not identify.
For a real fish, then, length error falls with distance rather than rising, because the body-depth shortfall shrinks with range faster than triangulation noise grows.

Every real fish has body depth, so we recommend working beyond 1~m.
None of the paired stereo frames were taken closer than 1~m, so the agreement of Section~\ref{sec:stereo_results} does not depend on that range.

\subsection{The field cohort}
\label{sec:field_results}

\begin{figure}
    \centering
    \includegraphics[width=0.6\columnwidth]{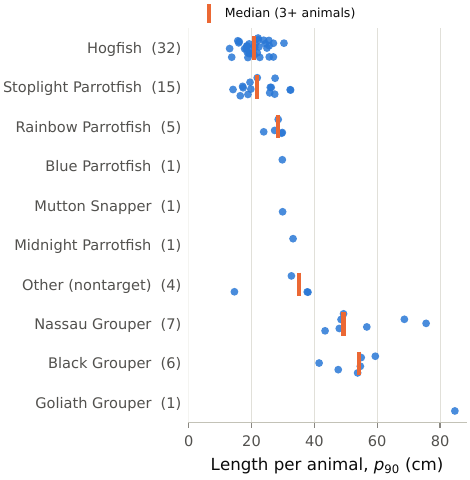}
    \caption{Length per animal by species over the seven dives, one point per animal at its $p_{90}$. The bar marks the median for species with three or more animals.}
    \Description{Strip plot of per-animal length grouped by species.}
    \label{fig:field_species}
\end{figure}

These open-water dives produced what Section~\ref{sec:introduction} argues a stock assessment needs, a length paired with a species for each individual fish.
Over seven dives, divers measured 73 wild fish from nine named species and a nontarget category, ranging from 13.0 to 84.7~cm with a median of 25.5~cm (Figure~\ref{fig:field_species}).

\begin{figure}
    \centering
    \includegraphics[width=0.7\columnwidth]{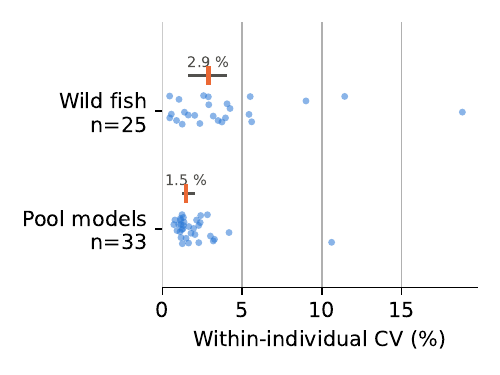}
    \caption{Within-individual repeatability, wild fish against posed models. Each point is one wild individual with three or more frames, or one target photographed repeatedly in one pool session. Bars give the median and its 95\% confidence interval.}
    \Description{Strip plot of within-individual coefficient of variation for wild fish and pool targets.}
    \label{fig:field_repeatability}
\end{figure}

The field cohort has no known lengths, so it cannot measure accuracy, but it can measure repeatability, which needs no reference.
We measure repeatability as the within-fish coefficient of variation, the spread of one fish's frames relative to their mean.
A calibration error is common to every frame of a fish, so it cancels out in this measure; therefore, it cannot detect a calibration shift shared by every fish on a dive, a term worth 2.4\% standard deviation in the pool (Table~\ref{tab:error_budget}).
For the 25 wild fish photographed in three or more frames, the median coefficient of variation is 2.9\% (95\% confidence interval 1.6 to 4.1\%).
For a pool target photographed repeatedly within one session, the median over 33 such cases is 1.5\% (95\% confidence interval 1.3 to 2.0\%), as Figure~\ref{fig:field_repeatability} shows.
A wild fish therefore varies about twice as much between frames as a posed model, though with 25 animals the two intervals overlap at their edges.
The likely sources are conditions the pool lacks, since the animal moves between frames, the water is turbid, and the snout and fork are harder to place.

Three properties of the cohort limit the precision of its lengths.
Each wild animal carries at most eight frames, so its $p_{90}$ is its longest, least foreshortened frame; in the pool, estimates from so few frames read slightly short, by 1.4\% at two frames (Section~\ref{sec:pool_results}).
Four of the seven field calibrations rest on only two or three usable slate frames, enough to span range but too few for the fit to check itself against, though dot noise alone puts at most 1.2\% uncertainty on their lengths at 3~m.
And 35 of the 162 measurements were taken closer than 1~m, where body depth makes them read several percent short (Section~\ref{sec:range_results}).

\section{Discussion}\label{sec:discussion}

The error sources measured in the pool are each small, and together they keep a pool measurement within its budget.
Table~\ref{tab:error_budget} collects them.
The terms that vary at random from one session, dive, fish, or frame combine to a standard deviation of 3.6\%, and two more only ever shorten a length: foreshortening, which the $p_{90}$ offsets by favoring the least foreshortened frames, and body depth, which nothing in the current pipeline offsets.
Against stereo imagery, the spread across seven fish was 10.1\%, which leaves about 9 points of standard deviation that no source in Table~\ref{tab:error_budget} explains.
Calibration is not the cause, since a calibration error shifts every fish on a dive alike, while the spread lies among fish on the same dive (Section~\ref{sec:stereo_results}); the remainder lies in sources that differ fish by fish, and separating them will take a larger paired study.

\begin{table}[]
    \centering
    \begin{tabular}{|l|r|l|l|}
        \hline
        Error source & Size (\% of length) & Behavior & Section \\
        \hline\hline
        Calibration scale between sessions & 2.4 (SD) & random per session & \ref{sec:pose_results} \\\hline
        Laser movement within a dive (two dives) & 2.2 (SD) & random per dive & \ref{sec:calibration_results} \\\hline
        Labeling noise at 2~m & 1.0 (SD) & random per frame & \ref{sec:range_results} \\\hline
        Few frames per fish & 1.3 (SD) & random per fish & \ref{sec:pool_results} \\\hline
        Pose within $15^\circ$ & up to 3.4 & always shortens & \ref{sec:pose_results} \\\hline
        Body depth & 1.5 at 2~m, 5.0 at 0.58~m & always shortens & \ref{sec:pool_results}, \ref{sec:range_results} \\\hline
    \end{tabular}
    \caption{The error sources measured in this paper, as magnitudes in percent of length. Each random term is a standard deviation: calibration scale, the spread of the scale errors of the five angle-experiment sessions ($-3.0$, $+0.2$, $-6.5$, $-2.3$ and $-3.3\%$; Section~\ref{sec:pose_results}); laser movement within a dive, the largest change observed between a dive's two ends, 3.7\%, treated as the edge of a uniform distribution; labeling noise, the 0.9~px dot and 2.3~px endpoint errors of Section~\ref{sec:range_results} for a 30~cm fish at 2~m; and few frames per fish, the spread a three-frame $p_{90}$ adds over an all-frames $p_{90}$ across the pool targets of Section~\ref{sec:pool_results}. These combine as independent errors, as the square root of the sum of their squares, to 3.6\%. The last two rows always shorten a length, so they add rather than cancel.}
    \label{tab:error_budget}
\end{table}

Two of these terms point at the design rather than the diver.
The mount is the largest source of error a design can remove, since its joints move the beam between dives by more than the budget allows (Section~\ref{sec:calibration_results}), and a stiffer mount would shrink the movement that per-dive calibration exists to catch.
Calibration is the one step divers must repeat on every dive, and the manual's closing calibration was the part most often left out, with only two of the seven field dives calibrated at both ends (Section~\ref{sec:field_cohort}), so it is the part of the protocol most worth making easier.
Body depth, the one bias left uncorrected, could be removed with a per-species correction wherever depth-to-length ratios are known.

These results have three limits.
The comparison with stereo imagery covers seven fish on one day, too few to bound error for a single wild fish, so the pool provides per-measurement accuracy evidence and the field cohort shows repeatability rather than accuracy.
Laser movement within the camera--laser plane during a dive cannot be seen in the photographs, and was measured on two field dives.
REEF briefed its divers in person, so the deployment shows that a citizen-science program can teach the system without its builders, but not yet that the written manual alone is enough.

\section{Conclusion}\label{sec:conclusion}

\fsl gives a citizen-science program fish lengths, paired with species, from commodity hardware and a single photograph.
On targets of known length, it recovers each to within 1.3\%, and on live fish it agrees with stereo imagery without detectable bias.
From the error sources these results measure, we derive a field protocol, in which the diver calibrates against a dive slate on every dive, photographs fish within $15^\circ$ of perpendicular and beyond 1~m, and takes at least ten frames of each.
REEF, trained once by the system’s builders, then taught its own divers and collected lengths paired with species for 73 wild fish across seven dives.

The design meets the \emph{commodity components} and \emph{deferred judgment} design goals set in Section~\ref{sec:introduction}, since every part is owned, ordered from a consumer retailer or printed, and species and length are judged afterward by people equipped to make them.
It meets \emph{manual-only training} in content, since everything a diver must know fits in a written manual, though in this deployment divers were also briefed in person.
That combination lets length paired with species come from divers wherever they already dive, adding to the existing surveys rather than replacing them.

\begin{acks}
\paragraph{Funding and resources.} This work was funded in part by the NSF Research Experiences for Undergraduates (REU) Site: Engineers for Exploration at UC San Diego (Award 2244123) and the Matthew Henson Fellowship Program supported by the UC HBCU Program. We also acknowledge the Computer Science and Engineering department at UC San Diego and the UCSD IEEE Supercomputing Club at San Diego Supercomputer Center for hosting compute resources for this research. We thank the REEF volunteer divers who collected the field data.

\paragraph{Competing interests.} Christopher L. Crutchfield, Brice X. Semmens,
and Ryan Kastner are co-founders of Aqua3D, a company that may commercialize
technology described in this paper. The measurement method is the subject of
a pending patent application~\cite{Kastner2025}.
\end{acks}

\bibliographystyle{ACM-Reference-Format}
\bibliography{refs}

\end{document}